\documentclass[10pt,twocolumn,letterpaper]{article}
\usepackage[accsupp]{axessibility}
\usepackage{lineno}
\usepackage[pagenumbers]{cvpr} 

\newcommand{\myparagraph}[1]{\noindent\textbf{#1}}

\definecolor{cvprblue}{rgb}{0.21,0.49,0.74}
\usepackage[pagebackref,breaklinks,colorlinks,allcolors=cvprblue]{hyperref}

 \usepackage{multirow} 
 \usepackage{courier}
 \usepackage{amsfonts}
 \usepackage{pifont}
 \usepackage{multibib}

\usepackage[toc,page,header]{appendix}
\usepackage{minitoc}

\newcites{sup}{References for Supplementary Material}
 
\def\paperID{5521} 
\def\confName{CVPR}
\def\confYear{2025}

\newcommand{\modelname}{GenieRedux}
\newcommand{\modelnamegt}{GenieRedux-G}

\newcommand{\randomtestname}{Basic Test Set}
\newcommand{\ppotestname}{Diverse Test Set}
\newcommand{\retroset}{\texttt{RetroAct}}

\title{Exploration-Driven Generative Interactive Environments}

\author{Nedko Savov\textsuperscript{1,\dag} \quad\quad Naser Kazemi\textsuperscript{1} \quad\quad Mohammad Mahdi\textsuperscript{1} \quad\quad Danda Pani Paudel\textsuperscript{1} \\ Xi Wang\textsuperscript{1,2,3} \quad\quad Luc Van Gool\textsuperscript{1} \\
\textsuperscript{1} INSAIT, Sofia University "St. Kliment Ohridski" \quad\quad \textsuperscript{2} ETH Zurich \quad\quad \textsuperscript{3} TU Munich
}

\begin{document}
\doparttoc 
\faketableofcontents 

\maketitle

\renewcommand{\thefootnote}{} 
\footnote{\dag Corresponding author. \textit{firstname.secondname}@insait.ai}

\begin{abstract}

    Modern world models require costly and time-consuming collection of large video datasets with action demonstrations by people or by environment-specific agents. To simplify training, we focus on using many virtual environments for inexpensive, automatically collected interaction data. Genie \cite{bruce2024genie}, a recent multi-environment world model, demonstrates
    simulation abilities of many environments with shared behavior. Unfortunately, training their model requires expensive demonstrations. Therefore, we propose a training framework merely using a random agent in virtual environments. While the model trained in this manner exhibits good controls, it is limited by the random exploration possibilities. To address this limitation, we propose AutoExplore Agent - an exploration agent that entirely relies on the uncertainty of the world model, delivering diverse data from which it can learn the best. Our agent is fully independent of environment-specific rewards and thus adapts easily to new environments. With this approach, the pretrained multi-environment model can quickly adapt to new environments achieving video fidelity and controllability improvement.

In order to obtain automatically large-scale interaction datasets for pretraining, we group environments with similar behavior and controls. 
    To this end, we annotate the behavior and controls of 974 virtual environments - a dataset that we name \retroset. 
    For building our model, we first create an open implementation of Genie - \modelname\ and apply enhancements and adaptations in our version \modelnamegt.  Our code and data are available at \href{https://github.com/insait-institute/GenieRedux}{https://github.com/insait-institute/GenieRedux}.
    
\end{abstract}

\section{Introduction}

Learning from interactive environments allows us to understand and represent the rules, the possible actions, and the consequences that govern them. As an alternative to laboriously hand-coded synthetic simulators, world models have emerged as deep learning tools for realistic environment modeling entirely from observations, commonly images of the observed environment \cite{menapace2021playable, yang2023unisim, alonso2024diffusion, bruce2024genie}.

\begin{figure}[t!]
    \centering
    \includegraphics[width=0.48\textwidth]{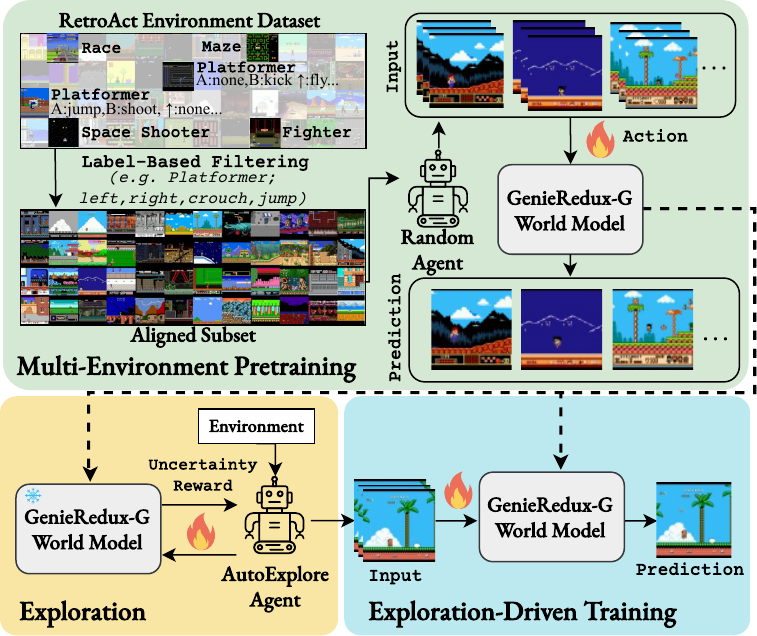}
    \vspace{-2mm}
    \caption{\textbf{Our proposed world model training framework.} It consists of a pretrained multi-environment world model on random agent data, and a new AutoExplore Agent that explores an environment and delivers diverse data for fine-tuning. }
    \label{fig:teaser}
    \vspace{-5mm}
\end{figure}

Previous work such as \cite{ha2018world, hafnermastering, lenz2015deepmpc} uses light world models to support goal-driven agents with goal-specific state representations. The focus is on coarse future predictions, not on their high visual quality. 
In contrast, the objective of recent world models is to achieve high-quality future predictions given past observations and actions. Such recent models are able to offer realistic action execution and even real-time interaction with people \cite{valevski2024diffusion, alonso2024diffusion}. This has become possible with the rise of diffusion, transformers \cite{dosovitskiy2020image, vaswani2017attention}, and state space models \cite{gu2023mamba}, and by borrowing architectural choices from video generation pipelines \cite{villegas2022phenaki,rombach2022high}. Typically, these generative models are designed to closely match a single selected environment. One of the state-of-the-art models, Genie, distinguishes itself by being trained on many visually diverse environments with similar dynamics, thus demonstrating generalization across new visuals. 

Building these high-quality statistical simulators requires diverse observations of the environment as well as of the actions to simulate. Some obtain this data by costly video dataset collection and curation with human demonstrations of the actions \cite{yang2023learning, bruce2024genie,alonso2024diffusion}. If actions are unavailable, an extra component is designed to predict them, which can introduce uncertainty compared to ground truth labels \cite{bruce2024genie, menapace2021playable}. Extension to new environments with new types of actions in this setting is difficult as it requires again an expensive data collection process. Others, such as \cite{valevski2024diffusion} have explored retrieving data with an environment-specific agent, in their case - the game Doom.


In this work, we propose a framework for accessible and effort-free training of world models in multiple environments. To this end, we first build \retroset - an annotated and curated large dataset of retro game environments (based on the environments of Stable Retro \cite{stable-retro}). 
We group them based on behavior labels and control descriptions. This grouping allows us to generate large-scale interaction datasets across environments with similar behaviors. Next, we pretrain a multi-environment world model \modelname\  - our open implementation of Genie \cite{bruce2024genie}, using a random agent. Unlike \cite{zhangprelar}, which reports the agent's improved behavior from pretraining, we aim to improve the world model. For this, we adapt \modelname\ to virtual environments and implement architectural and training procedure enhancements, resulting in the \modelnamegt\ model. We observe that just by training \modelnamegt\ on random interactions from subsets of 200 environments and 50 environments with mapped controls, automatically collected from \retroset, we are able to obtain control behavior (0.450 $\Delta$PSNR in 50 environments) and reasonable visual fidelity (26.36 PSNR in 50 environments).

As random actions are limited in their ability to explore the environment, we develop a method to obtain more diverse interaction data to improve the control behavior and visual fidelity of our model. To this end, inspired by \cite{sekar2020planning}, we develop our own environment-independent reward function, allowing an agent to explore different environments, entirely without relying on predefined environment rewards. While they aim at a high-performing goal-driven agent, we base our design on improving the underlying large world model for the simulation of environments in terms of higher visual fidelity and improved controllability. For graphical illustrations, see Fig.~\ref{fig:teaser}.
The objective of our exploration-driven agent is to maximize the world model's uncertainty, estimated by the classification entropy available in the observation prediction stage of \modelnamegt. Once the diverse data is obtained, we fine-tune \modelnamegt.
We show that this method leads to significant visual (up to 7.4 PSNR) and control (up to 1.4 $\Delta$PSNR) improvements, compared to random agent pretraining.

\begin{figure*}[h]
    \centering
    \includegraphics[width=\textwidth]{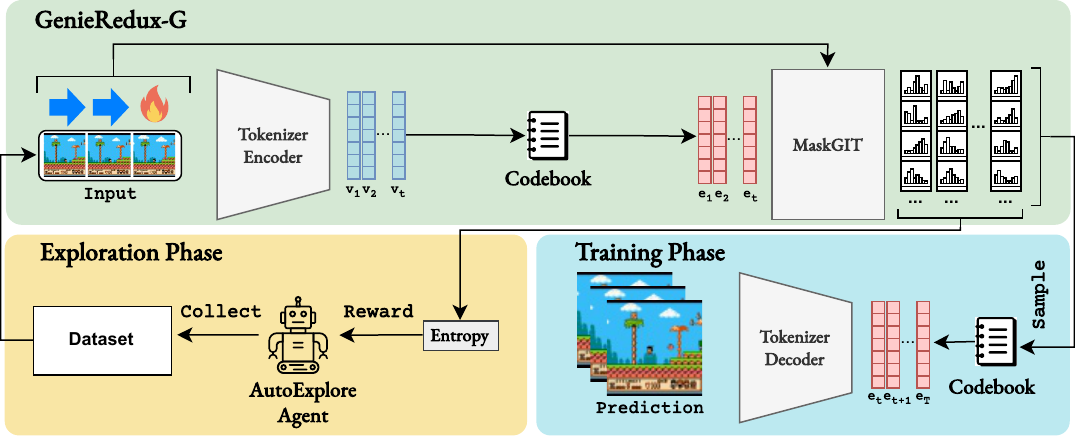}
    \vspace{-5mm}
    \caption{\textbf{Method Overview. }We propose an alternative to costly human interaction data collection - by exploring environments with an agent. The reward is solely based on the classification uncertainty of our model.}
    \vspace{-5mm}
    \label{fig:main_figure}
\end{figure*}

Our contributions are as follows:
\begin{itemize}
    \item A framework for training world models with cheap data collection - by training an exploration agent based on our world's model uncertainty.
    \item The implementation and release of \modelname\ and \modelnamegt\ - open Pytorch models based on \cite{bruce2024genie}. 
    \item Architectural and loss changes to the model leading to fidelity improvements, based on our tokenizer representation study.
    \item Preparing a large scale environment dataset for multi-environment world model training. 
\end{itemize}

\section{Related Work}
\myparagraph{World models.} Initially built as rough imagination models assisting reinforcement learning (RL) agents \cite{chiappa2017recurrent, ha2018world, hafner2019learning, hafner2023mastering, sekar2020planning}, world models have evolved into independent realistic video generation models conditioned on actions \cite{micheli2022transformers, chen2022transdreamer, yang2024video, robine2023transformer}. They facilitate task-specific agent training by providing predictive representations of the environment. Inspired by \cite{ha2017draw}, \citet{ha2018recurrent} use a VAE to encode visual observations into latent states, with an MDN-RNN predicting future states based on prior states, actions, and VAE outputs to facilitate policy learning. DreamerV2 \cite{hafner2020mastering} introduce an RL agent, achieving human-level performance in Atari. It encodes images with a CNN and computes posterior and prior stochastic states using recurrent states. Unlike our work, though, it does not assess the agent's impact on world model improvement nor generalize task rewards across different environments.

World models also aim to generate realistic video conditioned on actions \cite{menapace2021playable, yang2023unisim, hu2023gaia}. Genie \cite{bruce2024genie} trains a video tokenizer and a Latent Action Model (LAM) for dynamic next-frame generation. GAIA-1 \cite{hu2023gaia} tackle autonomous driving in unstructured settings by encoding multi-modal inputs into a unified representation and predicting image tokens based on prior inputs, using an autoregressive transformer. \citet{menapace2021playable} employ an encoder-decoder architecture in which the predicted action labels act as a bottleneck, allowing a user to control the generated video by a discrete action. The key gap in these works is automatic data collection, which is addressed in our approach.

\noindent \textbf{Efficient exploration.} The importance of efficient exploration in RL is highlighted by \cite{kakade2002approximately}. Early methods enhanced exploration by adding noise~\cite{lowe2017multi, fujimoto2018addressing} or using entropy regularization~\cite{mnih2016asynchronous}, but they have action space limitations and often fail with complex dynamics, where varied actions do not always drive meaningful exploration. A more direct approach uses heterogeneous actors~\cite{horgan2018distributed, schaul2015prioritized, kapturowski2018recurrent} with diverse exploration strategies to enhance environment exploration. Bayesian methods~\cite{srinivas2009gaussian, thompson1933likelihood} have also been introduced to create acquisition functions for uncertainty-driven exploration~\cite{osband2016generalization, azizzadenesheli2018efficient, metelli2019propagating, osband2015bootstrapped, osband2016deep, osband2018randomized}, but often struggle to generalize to high-dimensional inputs like images.

Recent exploration methods emphasize state novelty
\cite{tang2017exploration, ostrovski2017count, bellemare2016unifying, martin2017count, machado2020count, choshen2018dora, burda2018exploration, zhang2021made}, focusing on encouraging agents to assess novelty only after visiting states. In contrast, our approach, inspired by \cite{burda2018exploration, pathak2019self, sekar2020planning}, uses model disagreement to proactively guide agents to states with the highest potential without environment target-driven reward. \cite{burda2018large,pathak2017curiosity} propose exploration agents driven by uncertainty in state transition and simple feature extractors. Instead, we propose an exploration agent designed to improve a world model that does not model states. Plan2Explore~\cite{sekar2020planning} enables agents to seek novel states using a reward that maximizes the state entropy of an RSSM model. While Plan2Explore improves goal-driven agents with their framework, we improve a modern transformer world model with a novel exploration-based reward using token uncertainty.

Rather than relying on world models, EX2~\cite{fu2017ex2} learns a classifier to distinguish visited states, providing intrinsic rewards for states that are difficult for the classifier to differentiate. KL-divergence-based approaches~\cite{kim2019curiosity, klissarov2019variational, lee2019efficient}, guide exploration by comparing distributions. For example, SMM~\cite{lee2019efficient} computes the KL divergence between the policy-induced state distribution and a uniform target. ~\citet{tao2020novelty} propose an intrinsic reward based on the distance between a state and its nearest neighbors in a low-dimensional feature space. However, low-dimensionality leads to information loss, restricting full state space exploration — an issue we address by the use of a world model.

\begin{figure}[t]
    \centering
    \includegraphics[width=0.48\textwidth]{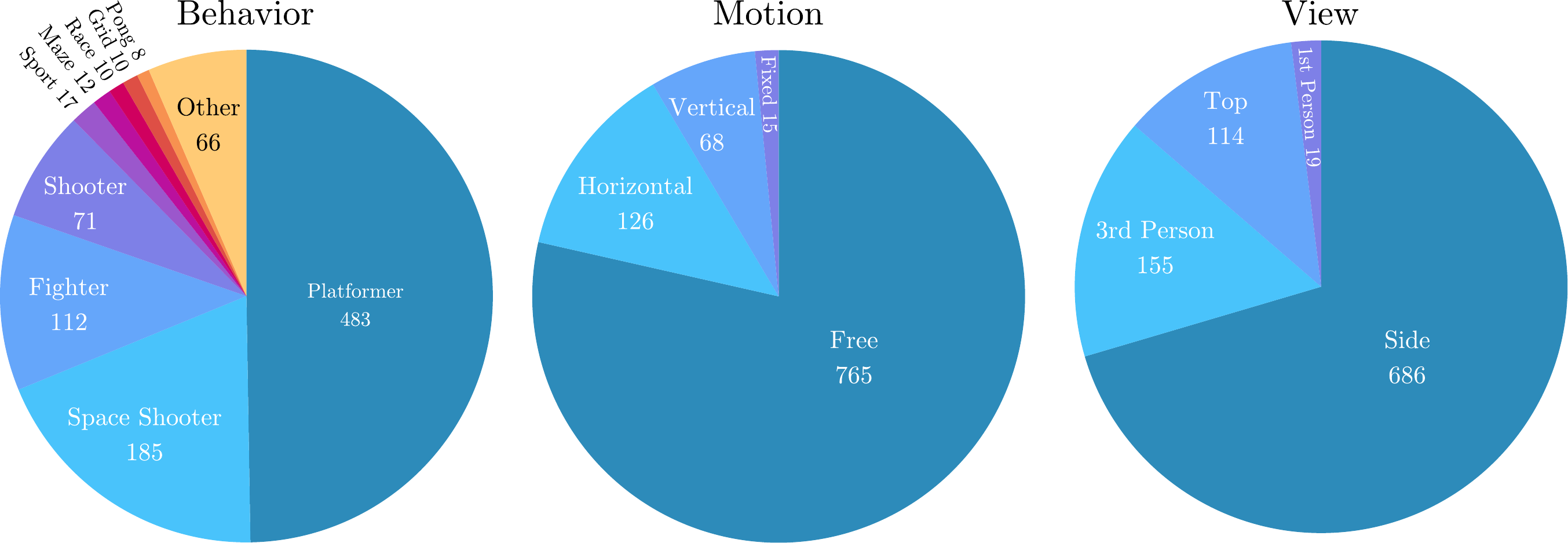}
    \vspace{-4mm}
    \caption{\textbf{\retroset\ Annotation.} Description of environments in \retroset\ by annotated attribute. Better viewed zoomed.}
    \label{fig:data_visualization}
    \vspace{-6mm}
\end{figure}

\begin{table*}[h]
    \centering
    \caption{\textbf{Comparison of \retroset\ dataset to others.}}
    \vspace{-3mm}
    \begin{tabular}{c|cccccccc }
        \hline
        \noalign{\vspace{0.1em}}
        \textbf{Dataset} & \textbf{Type} & \textbf{\#Environments} & \textbf{\shortstack{Diverse \\ Behaviors}} & \textbf{Open} & \textbf{\shortstack{Behavior \\ Annotation}} & \textbf{\shortstack{Control \\ Annotation}} \\
        \hline
        \texttt{Coinrun} \cite{cobbe2019quantifying} & Environments & 1 & \ding{55} & \checkmark & \ding{55} & \ding{55} \\
        \texttt{ALE} \cite{bellemare13arcade} & Environments & 57 & \checkmark & \checkmark  & \ding{55} & \ding{55} \\
        \texttt{Stable Retro} \cite{stable-retro} & Environments & 1003 & \checkmark & \checkmark  & \ding{55} & \ding{55} \\
        \texttt{Platformers} \cite{bruce2024genie} & Videos & Unknown & \checkmark & \ding{55} & \ding{55} & \ding{55}\\
        \retroset (Ours) & Environments & 974 & \checkmark  & \checkmark & \checkmark & \checkmark \\
        \hline
    \end{tabular}
    \vspace{-5mm}
    \label{tab:retroact_comparison}
\end{table*}

\section{RetroAct Dataset}
We first tackle the problem of accessible training of multi-environment world models by building a framework for cheaply acquiring multi-environment interaction data. In particular, we aim to collect interactions of similar actions in many environments. Instead of relying on expensive human interaction, we obtain and curate a collection of virtual environments. As a source, we use the Stable Retro framework by \cite{stable-retro}, which is a collection of retro games across multiple platforms, with an accompanying starting state. We make no use of the defined rewards. We obtain almost all the supported games (974).

This raw set contains an environment mix of very different visuals and behaviors. However, in our setting of learning from similar dynamics, it is required to establish correspondence between the environments' behaviors. We perform annotation where for each environment three aspects are classified. The motion style classifies the general style of what and how is moved by the controls, closely relating to game genre; the camera viewpoint; the control axis describing in which direction the player can be moved. The label distributions are shown in Fig. \ref{fig:data_visualization}. In Tab. \ref{tab:retroact_comparison} we compare our \retroset\ with other related datasets. \retroset\ distinguishes itself by providing behavior and control annotations, while maintaining a high number of environments.

It is discovered that the most prevalent type of environment in our set is the platformer - 483 titles. As the largest subset, we filter only these games for further use, as it is required to have many environments exhibiting similar controls.
Five motion actions are defined for our model - moving \texttt{left}, \texttt{right}, \texttt{up}, \texttt{down} and \texttt{jump}. Each game has its own mapping of buttons to actions.
Therefore, we generate a short clip of each of the 5 selected actions for each of the 483 titles and build an annotation tool to observe and annotate the executed action. Eventually, we annotated 2,925 behavior and 2,898 control labels.

After experimenting, we observed that models require more training with a higher number of environments, so we defined two subsets of our large set to handle computational cost: a subset consisting of the first 200 games of 483 behavior-filtered games for pretraining, and another subset of 50 randomly
selected action-consistent games using RetroAct’s action labels for fine-tuning.

We collect a large scale dataset by launching a random agent in all of the environments, collecting actions and observations. From the 200-game set we build \texttt{Platformers-200} - a dataset with 10,000 episodes (50 episodes per game) with 500 frames each at most, resulting in ~4.6mln images. From the 50-game set we obtain \texttt{Platformers-50} - 5000 episodes (100 per game) of length at most 1000, resulting in ~4.8mln images. In our protocol, we take 1\% of the sessions of each environment as a validation set. We show that using a random agent is already sufficient to learn a level of controllability and later build on top with an exploration agent of our design.

To validate our GenieRedux implementation, we implement the CoinRun case study in \cite{bruce2024genie}. Using the protocol from above, we obtain a dataset of 10k episodes with a maximum length of 500, resulting in 4mln images.

\section{Multi-Environment World Model}

 Given virtual environments, our first goal is to automatically obtain a dataset of image sequences $I_1, ..., I_N$ and corresponding actions $a_1, ..., a_{N-1}$. Given a sequence $I_1, ..., I_N$ and past and future actions $a_1, ..., a_{N+T-1}$, our world model aims to predict the future $T$ frames $I_{N+1}, ..., I_T$, corresponding to the actions executed. 

\paragraph{\modelname.} As Genie \cite{bruce2024genie} is not made available by the authors, we create an open implementation and call it \modelname. We validate our implementation quantitatively and qualitatively in Sec. \ref{sec:experiments} and Sup.Mat F.
It consists of three components. A video \textbf{Tokenizer} encodes input frame sequences into spatio-temporal tokens:
$e_1,...,e_N=T_{enc}(I_1,...,I_N)$, and decodes back to images: $I_1,...,I_N=T_{dec}(e_1,...,e_N)$.
A \textbf{Latent Action Model} encodes input frame sequences into spatio-temporal tokens: $a_1,...,a_{N-1} = LAM_{enc}(I_1,...,I_{N-1})$, and decodes them to reconstruct future prediction $I_2,...,I_{N} = LAM_{dec}(a_1,...,a_{N-1})$. A \textbf{ Dynamics module} predicts the next frames based on partially masked frame tokens and actions: $I_{2},...,I_{N+T-1}=D(e_1,...,e_N, ...,e_{N+T-1};a_1,...,a_{N+T-1})$, where in inference $e_N,..,e_{N+T-1}$ are masked. We adhere closely to Genie's specifications for implementing these components.

All components use the causal Spatial Temporal Transformer (STTN)  \cite{xu2020spatial}.
We use Position Encoding Generator (PEG) \cite{chu2021conditional} for spatial and temporal attention, and Attention with Linear Biases (ALiBi) \cite{press2021train} for temporal attention.

We train our models with a sequence size of 16 frames and resolution of 64x64 to address computational limitations. We train a U-Net-based superresolution network on 50K data samples to upscale the output to 256x256. (Sup.Mat. B)

 \paragraph{\modelnamegt.} Building upon the base model, we offer a variant - \modelnamegt, which is adapted to virtual environments and contains architectural and training improvements. While \modelname\ uses an indispensable LAM model to obtain the actions, we discard it, as ground truth actions are available from our agent. Instead, the one-hot actions are concatenated to each layer of the Dynamics module for conditioning. In this way, we avoid the uncertainty of a prediction.

The Dynamics module consists of an ST-ViViT encoder, followed by a MaskGIT architecture \cite{chang2022maskgit}, which predicts indices from the tokenizer's codebook for randomly masked input tokens during training, according to a schedule. As standard cross-entropy is used, token classification has the drawback to penalize equally any prediction different from the ground truth. However, close tokens in the codebook result in significantly fewer changes than far tokens, as also shown in Sec. \ref{par:exp_tokenizer_repr}. To enable this concept of a distance between tokens in the classification of $N_E$ tokens, we design a Token Distance Cross-Entropy (TDCE) Loss:

\vspace{-4mm}
\begin{align}TDCE(x,y)=(y^TK)\cdot softmax(x) + CE(x,y)
\end{align}

Here $x\in\mathcal{R}^{N_E}$ is the prediction logits, $y\in\mathcal{R}^{N_E}$ is the ground truth one-hot class. $K\in\mathcal{R}^{N_E\times N_E}$ is a precomputed table at the start of training of the cosine distances between all tokens
; $CE(.)$ denotes standard Cross-Entropy Loss.
When an incorrect token class is given probability, it is penalized based on its distance to the ground truth class.

MaskGIT's design is to take as input learnable embeddings, indexed by the tokens predicted by the Tokenizer. They are randomly initialized, and therefore contain none of the content of the tokens. Given that that the encoding itself and the distance between tokens can contribute to Dynamic module's performance, we add a skip connection by adding the embedding to the token itself, which improves visual fidelity and controllability of the model.

\paragraph{AutoExplore Agent} We extend our framework with an exploration agent that obtains data by going deeper into the environments. We name it AutoExplore Agent. The reward of the agent is entirely based on the world model performance and operates without any environment rewards. Therefore, it can be trained in various environments without tuning to their specifics or relying on a reward definition.

The design of our reward is based on the fact that \modelnamegt\ employs classification for token prediction. Each token is predicted by sampling from a categorical distribution over the codebook. We first obtain all $N_T$ token prediction distributions by running \modelnamegt-50 5 steps back from the current observation $I_c$ for which we want to estimate the reward. We provide 2 images $I_{c-4}, I_{c-3}$, predict 3 images - $I_{c-2}, ..., I_{c}$, and take the distributions of the predicted tokens of $I_c$ to obtain $x=[x_1, ..., x_t, ..., x_{N_T} ]$. We evaluate the uncertainty per predicted token $u_t$ by calculating the entropy over the categorical distribution and normalize it in the range $[0,2]$:

\vspace{-4mm}
\begin{align}
u_t=\frac{2\cdot \sum_i^{N_T} x_i\cdot log(x_i)}{N_e}
\end{align}

Studying the properties of the Tokenizer representation, we find that a prevalent token is learned representing static parts of the environment. Only the changing parts generate high uncertainty and, therefore, we take the subset $S_{top}$ of 25\% highest uncertainties of the entire set of uncertainties $S=\{u_t\}$. The reward, shown in Eq. \ref{eq:reward}, establishes the agent's goal to collect data that maximizes uncertainty of the world model.

\vspace{-4mm}
\begin{align}\label{eq:reward}
S_{25\%} = \{ u \in S \mid u \geq Q_{75}(S) \} \\
R(I_c)=\frac{1}{|S_{25\%}|} \sum_{u \in S_{25\%}}u
\end{align}

Our agent is an actor-critic, trained with the Policy Gradient method. For the agent architecture, we follow \cite{micheli2022transformers}. It consists of a CNN encoder followed by an LSTM. As standard in RL, 4 frames are stacked, max-pooled, and the result is the input to the agent for a single time step.

\paragraph{Exploration-driven World Model Training.}
We initially pretrain \modelnamegt\ on \texttt{Platformers-200} and fine-tune on \texttt{Platformers-50} to obtain the model \modelnamegt-50. Then, we train AutoExplore Agent by using \modelnamegt-50, using it as a source of reward.
The details of training the agent are presented in Sup.Mat A.3.

Running the trained exploratory agent on a selected environment, we obtain a new diverse dataset with action demonstrations under unseen scenes. We first fine-tune the decoder of the Tokenizer for 1,000 iterations to adapt to the new unseen scenes. The Dynamic module of \modelnamegt\ is then fine-tuned on the new data to achieve greater visual fidelity and controllability under new conditions. 
In order to build test sets to evaluate our approach, we train an Agent-57 model for each of the environments we explored, using the available environment rewards. More details on the test setup are provided in Sup.Mat A.2.

For visual fidelity evaluation, we use FID (Fréchet inception distance) \citet{heusel2017gans}, PSNR (signal-to-noise ratio) and SSIM (structural similarity index measure) \citet{Wang2004ImageQA}. To evaluate controllability, we use the recently proposed \(\Delta_t\text{PSNR}\) metric \cite{bruce2024genie}, which compares the visual effect of the ground truth action ($\hat{x}_t$) versus a random action ($\hat{x}_t'$):
\(\Delta_t \text{PSNR} = \text{PSNR}(x_t, \hat{x}_t) - \text{PSNR}(x_t, \hat{x}_t'),
\), where $x_t$ is the ground truth frame at time $t$. A higher $\Delta_t \text{PSNR}$ indicates a higher level of controllability. As in \citet{bruce2024genie}, for all experiments we report \(\Delta_t\text{PSNR}\) with \(t=4\).

\section{Experiments}\label{sec:experiments}
\paragraph{Comparing \modelname\ and \modelnamegt.}
We implement the original CoinRun case study with a random agent, as advised by \cite{bruce2024genie}, in order to validate and compare \modelname\ with LAM, and \modelnamegt\, which uses agent-provided actions instead. In this study, the presence of LAM is the only difference between the models. We first train on a dataset, collected by a random agent. Visual fidelity results are in Tab. \ref{tab:base_fidelity_eval}. Our \modelname\ implementation exhibits high visual quality and matches all seven CoinRun environment actions, as well as progressing environment motions (demonstrated in Sup.Mat. F). However, as demonstrated by the metrics, \modelnamegt\ shows superior visual fidelity and controllability (more in Sup.Mat. F), as it avoids the uncertainty of LAM prediction. This study demonstrates that even using a random agent can result in action performance abilities in the world model.

\begin{table}[t!] 
\centering

    \centering
    \caption{\textbf{Comparison of \modelname\ and \modelnamegt\ on Basic Test Set. } Peformed on a test set, collected from the Coinrun environment with randomly sampled actions.}
    \vspace{-3mm}
    \begin{tabular}{c|ccc}
      \hline
      \multirow{2}{*}{\textbf{Model}} & \multicolumn{3}{c}{\textbf{\randomtestname}} \\
       & \textbf{FID$\downarrow$} & \textbf{PSNR$\uparrow$} & \textbf{SSIM$\uparrow$} \\
      \toprule
      Tokenizer & 18.14 & 38.25 & 0.96 \\
      \midrule
      LAM & 37.01 & 33.97 & 0.92 \\ 
      \midrule
      \modelname & 21.88  & 25.51 & 0.77 \\ 
      \modelnamegt & \textbf{18.88} & \textbf{33.41} & \textbf{0.92} \\
      \bottomrule
    \end{tabular}
    \label{tab:base_fidelity_eval}
\end{table}
\begin{figure}[t!]
    \centering
    \captionof{table}{\textbf{Comparison of \modelname\ and \modelnamegt\ on Diverse Test Set.} The models are trained with data collected by random agent and trained agent (-TA), and tested on data collected by a trained agent from the Coinrun environment.}
    \vspace{-3mm}
    \begin{tabular}{c|cccc}
        \hline
        \multirow{2}{*}{\textbf{Model}} & \multicolumn{3}{c}{\textbf{\ppotestname}} \\
         & \textbf{FID$\downarrow$} & \textbf{PSNR$\uparrow$} & \textbf{SSIM$\uparrow$} \\
        \toprule
        Tokenizer & 19.13 & 35.85 & 0.94 \\
        Tokenizer-TA  & \textbf{11.63} & \textbf{40.62} & \textbf{0.97} \\
        \midrule
        \modelname & 23.97 & 23.82 & 0.73  \\
        \modelnamegt  & 19.51 & 31.66 & 0.90 \\
        \modelname-TA  & 12.57 & 31.97 & 0.90 \\
        \modelnamegt-TA  & \textbf{12.40} & \textbf{34.44} & \textbf{0.92} \\
        \bottomrule
    \end{tabular}
    \vspace{-5mm}
    \label{tab:main_eval}
\end{figure}

Next, we train an actor-critic agent with PPO on the environment reward, following \cite{cobbe2019quantifying} to collect data and train \modelname-TA and \modelnamegt-TA. Tab. \ref{tab:main_eval} shows evaluation on a test set collected by a trained agent. \modelnamegt\ outperforms \modelname\ on all settings. Furthermore, models trained on diverse agent-collected data are visually superior to those trained on random agents. The higher $\Delta$PSNR of 1.89 for \modelnamegt-TA compared to 0.70 for \modelnamegt\ shows the superiority of diverse data training in controllability. (more in Sup.Mat. F)

\begin{figure}[t]
    \centering
    \includegraphics[width=0.45\textwidth]{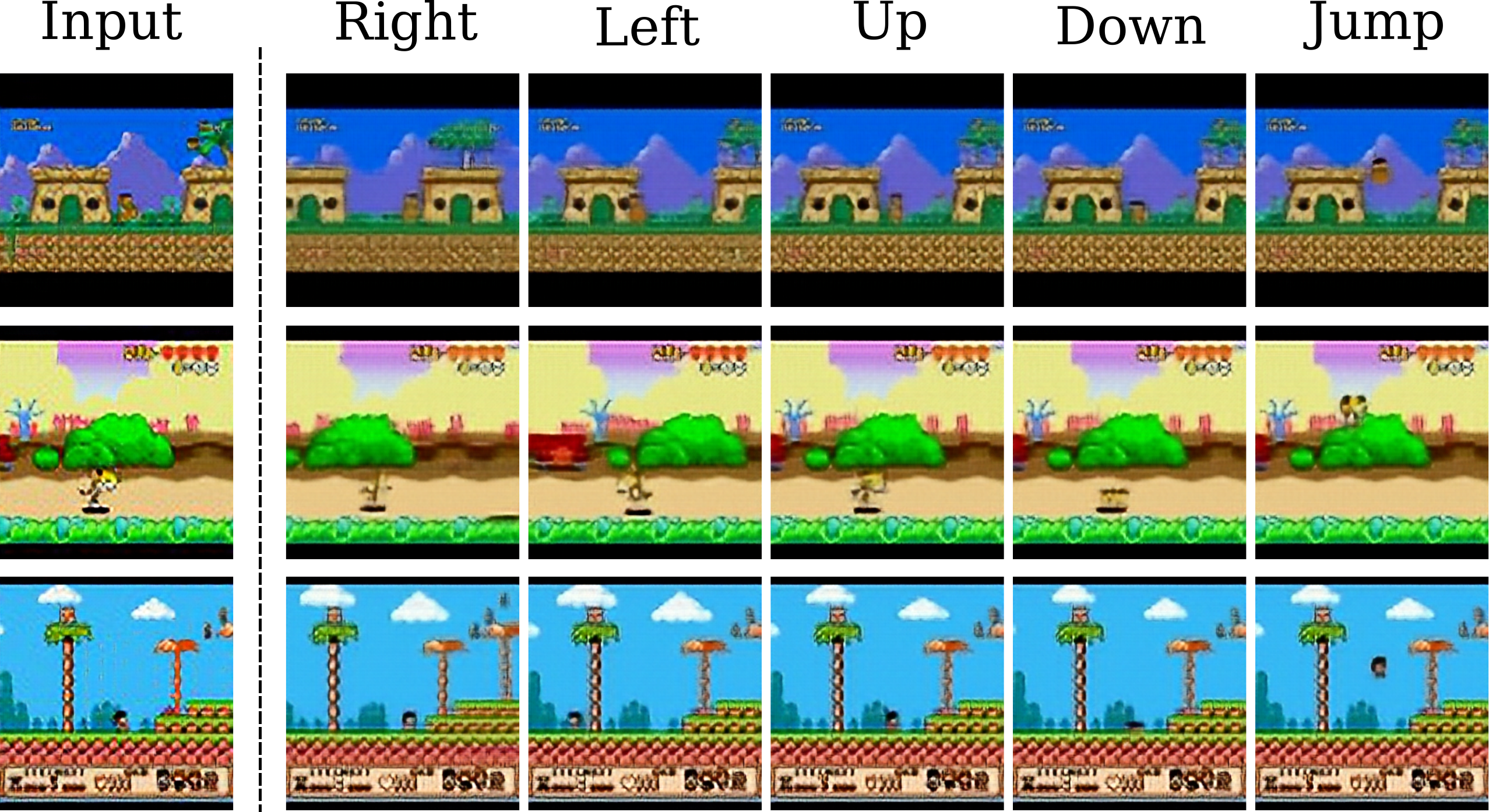}
    \vspace{-3mm}
    \caption{\textbf{Control Of \modelnamegt-50.} Demonstrating all controls of our multi-environment model on multiple games.}
    \vspace{-5mm}
    \label{fig:multienv_control}
\end{figure}

\paragraph{Multi-Environment Models.}
Here, we evaluate the models we initially train on many environments from \retroset. \modelnamegt-200\ is pretrained on the \texttt{Platformers-200} dataset for 180k iterations. On the validation set, we obtain 23.32 PSNR and 17.12 FID. Using this model as a base, \modelnamegt-50 is trained on \texttt{Platformers-50}. Its quantitative evaluation on a test set of 10k sessions separately generated from the selected 50 environments is at the start of Tab. \ref{tab:ablation}. As the 50 environments are selected with corresponding action controls between each other, we see a boost in the quality of prediction. Fig. \ref{fig:multienv_control} demonstrates that the instructed action is executed successfully by \modelnamegt. As the up action is rarely used, it serves more as a no-operation action. (more in Sup.Mat C.1)

\begin{table}[t]
    \centering
    \caption{\textbf{Ablation study on improvements in \modelnamegt.}}
    \vspace{-2mm}
    \begin{tabular}{c|ccc}
        \hline
        \textbf{Model} & \textbf{FID$\downarrow$} & \textbf{PSNR$\uparrow$} & \textbf{SSIM$\uparrow$} \\
        \hline
        \modelnamegt-200 & 22.31 & 25.11 & 0.80 \\
        \modelnamegt-50 & 23.80 & 26.36 & 0.84 \\
        + Token Input & 22.96 & 26.65 & 0.84 \\
        + TDCE Loss & 22.95 & 27.06 & 0.85 \\
        Autoregressive & \textbf{22.11} & \textbf{28.07} & \textbf{0.88}\\
        \hline
    \end{tabular}
    \vspace{-2mm}
    \label{tab:ablation}
\end{table}

\begin{table*}[t]
    \centering
    \caption{\textbf{Quantitative Results on 3 environments.} We evaluate the benefit of the data from the propose AutoExplore Agent to our models. \modelnamegt-50 is our pretrained world model on 50 environments. \modelnamegt-50-ft are fine-tuned models using data from a random agent or AutoExplore (Exploration). \modelnamegt\ denotes a non-fine-tuned model, trained with the exploration data. }
    \vspace{-2mm}

    \begin{tabular}{c|c|c|cccc}
        \hline
        \textbf{Environment} & \textbf{Strategy} & \textbf{Model} & \textbf{FID$\downarrow$} & \textbf{PSNR$\uparrow$} & \textbf{SSIM$\uparrow$} & \textbf{$\Delta$PSNR$\uparrow$} \\
        \hline
        & \multirow{2}{*}{Random} & \modelnamegt-50  & 41.99 & 26.32 & 0.81 & 0.83 \\
        & & \modelnamegt-50-ft & 42.34 & 27.04 & 0.81 & 1.19 \\
         \cline{2-7}
        \multirow{3}{*}{Adventure Island II} & \multirow{3}{*}{Exploration}  &
         Tokenizer-ft & 11.01 & 38.95 & 0.98 & - \\
        & &\modelnamegt\ & 11.94 & 28.33 & 0.88 & 0.37 \\ 
        & & \modelnamegt-50-ft & \textbf{12.77} & \textbf{30.60} & \textbf{0.90} & \textbf{1.47} \\
        
         \cline{2-7}
        & Random Autoregressive & \modelnamegt-50-ft  & 41.55 & 27.82 & 0.83 & 1.24 \\
        & Exploration Autoregressive& \modelnamegt-50-ft  & \textbf{11.33} & \textbf{33.61} & \textbf{0.94} & \textbf{2.09} \\
        \hline
        \multirow{7}{*}{Super Mario Bros} & \multirow{2}{*}{Random} & \modelnamegt-50 & 29.83 & 34.24 & 0.94 & \textbf{0.56} \\
        & & \modelnamegt-50-ft & 30.13 & 34.54 & 0.94 & \textbf{0.54} \\
        \cline{2-7}
         & \multirow{3}{*}{Exploration} & Tokenizer & 8.09 & 42.00 & 0.99 & - \\
        & & \modelnamegt\ & \textbf{9.56} & 34.00 & 0.95 & 0.09 \\
        & & \modelnamegt-50-ft & \textbf{9.55} & \textbf{36.13} & \textbf{0.97} & \textbf{0.57} \\
        \cline{2-7}
        & Random Autoregressive & \modelnamegt-50-ft & 30.84 & 34.85 & 0.95 & 0.57 \\
        & Exploration Autoregressive & \modelnamegt-50-ft & \textbf{9.33} & \textbf{37.77} & \textbf{0.97} & \textbf{0.76} \\
        \hline
        \multirow{7}{*}{Smurfs} & \multirow{2}{*}{Random} & \modelnamegt-50 & 79.51 & 21.47 & 0.69 & 0.47 \\
        & & \modelnamegt-50-ft & 80.61 & 21.83 & 0.70 & 0.65 \\
        \cline{2-7}
         & \multirow{3}{*}{Exploration} & Tokenizer & 17.86 & 35.61 & 0.98 & - \\
        & & \modelnamegt\ & 20.43 & 35.42 & 0.80 & 0.85 \\
        & & \modelnamegt-50-ft & \textbf{20.01} & \textbf{27.45} & \textbf{0.85} & \textbf{1.55} \\
        \cline{2-7}
        & Random Autoregressive & \modelnamegt-50-ft & 80.16 & 22.16 & 0.71 & 0.69 \\
        & Exploration Autoregressive & \modelnamegt-50-ft & \textbf{18.97} & \textbf{29.53} & \textbf{0.90} & \textbf{2.06} \\
        \hline
    \end{tabular}
    \vspace{-3mm}
    \label{tab:results}
\end{table*}

\paragraph{Ablation Study.}
In this experiment we evaluate the additive gain of each proposed improvement in \modelnamegt\ - the additive token input and training with the Token Distance Cross-Entropy Loss. The ablation is performed on a generated test set of 10k sessions, each 500 frames long. The data is collected using a random action policy from the environments in \texttt{Platformers-50}. Visual fidelity evaluation is provided in Tab. \ref{tab:ablation}. It can be seen that each component gives our model a benefit in terms of visual fidelity. Finally, we perform an autoregressive evaluation of the best model to achieve our highest score.

\begin{figure}[t]
    \centering
    \includegraphics[width=0.45\textwidth]{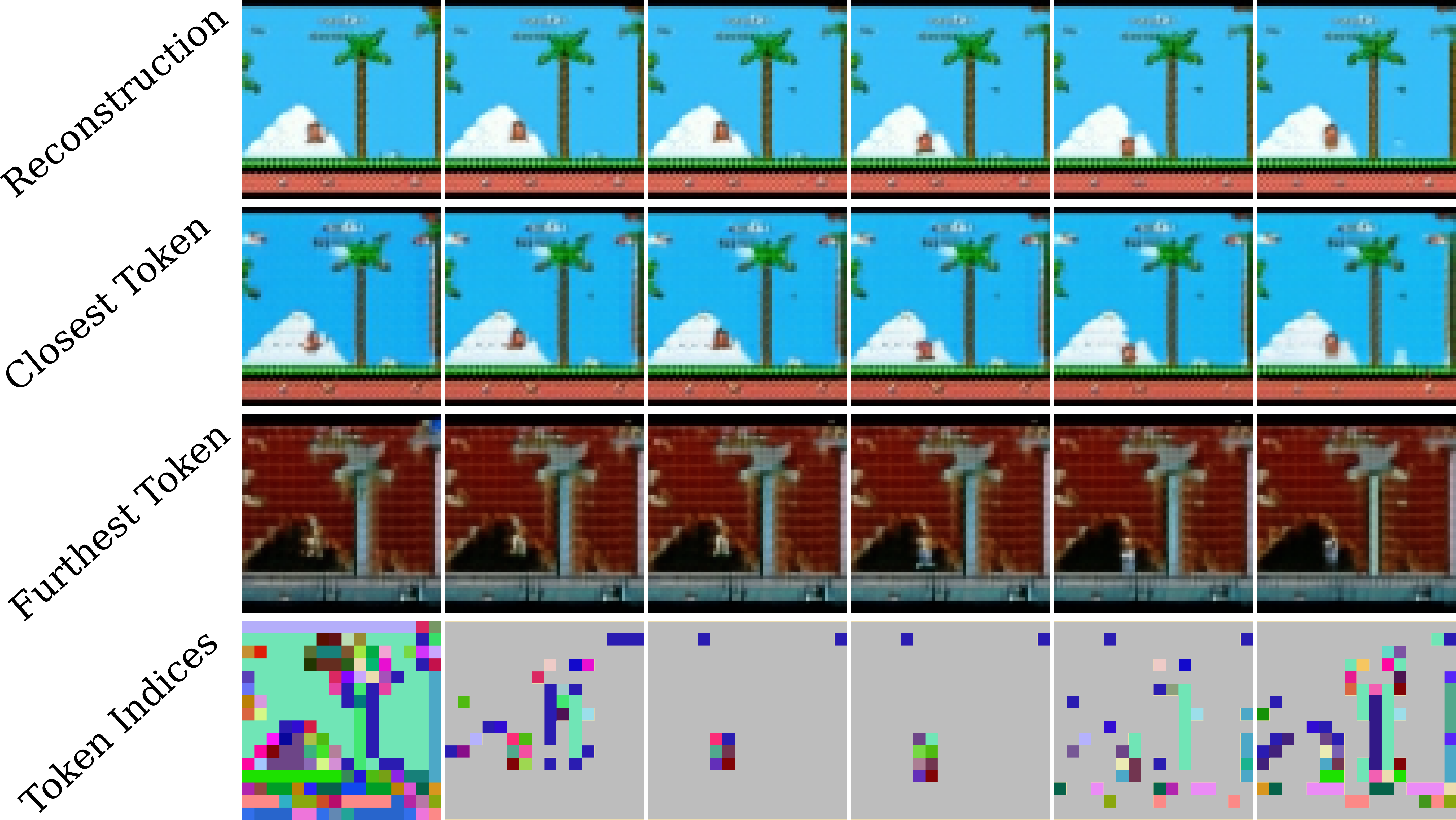}
    \vspace{-2mm}
    \caption{\textbf{Tokenizer Representation.} Reconstruction images from the tokenizer, and the effect of replacing each token with its closest and furthest in the codebook. Lastly, we visualize the indices of the predicted tokens.}
    \vspace{-5mm}
    \label{fig:tokenizer_representation}
\end{figure}

\paragraph{Tokenizer Representation Study.}\label{par:exp_tokenizer_repr}
This experiment provides insights into the inner workings of \modelnamegt\ to motivate our proposed changes.
As the Dynamics module operates entirely on the token representation, we examine it closely. 
Fig. \ref{fig:tokenizer_representation} shows the reconstructions of an input sequence (first row) and the visualized token representation (last row), where each predicted token index is assigned a different color. The visual features of the first frame are captured by various tokens.
Starting with the second frame, the representation drastically changes - a token is specialized in representing the static frame regions compared to the past, while all motion regions are updated with new content. Observing that visually similar patches predict identical or similar tokens, we replace each predicted token with its closest in the codebook. We only keep the special background token unchanged. In the second row of Fig. \ref{fig:tokenizer_representation} we show the resulting reconstruction - while some blurriness appears, the image remains largely the same. Conversely, replacing each token with its furthest in the codebook (third row) results in a significantly different image. This property - closer tokens having more similar appearance - motivates our Token Distance Cross-Entropy Loss, which penalizes predicting tokens further away from the ground truth.

\begin{figure}[t]
    \centering\includegraphics[width=0.45\textwidth]{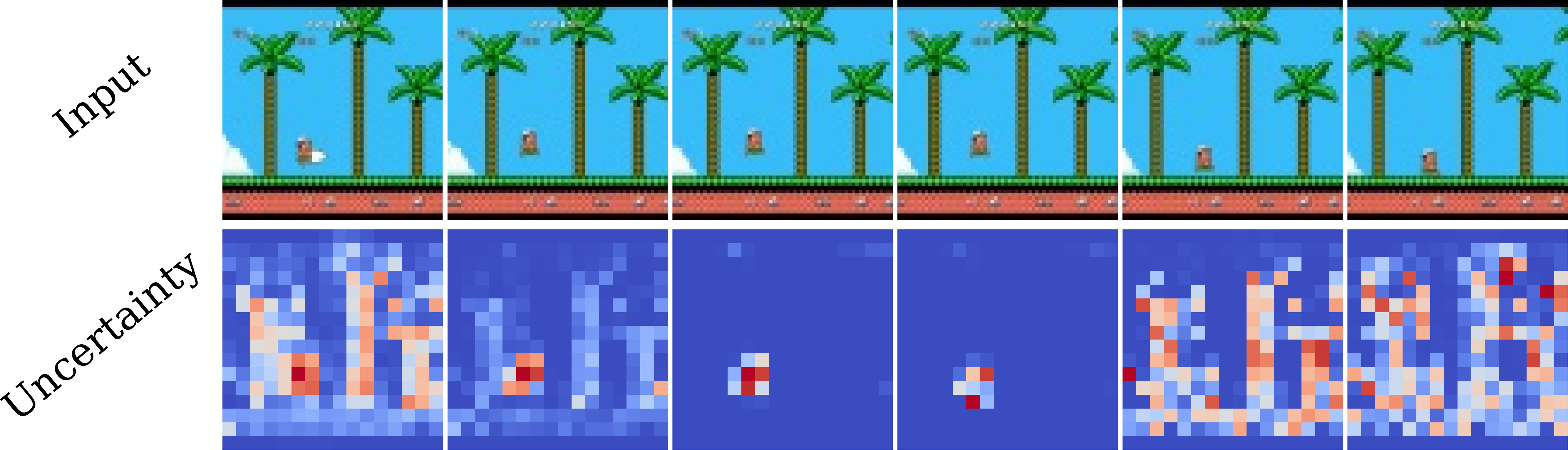}
    \vspace{-2mm}
    \caption{\textbf{Dynamics Uncertainty.} Shown is the uncertainty per token predicted for each image of an example sequence. Uncertainty is generated in the regions of motion.}
    \vspace{-5mm}
    \label{fig:dynamics_uncertainty}
\end{figure}

Fig. \ref{fig:dynamics_uncertainty} visualizes the uncertainty of \modelnamegt-50 for each predicted token of its Dynamics module given a sequence. The uncertainty metric is the entropy of the classification over 1024 codebook tokens. Tokens corresponding to motion have the highest uncertainty; other regions are mostly classified as the "static" token. Thus, minimal character movement yields low uncertainty, while forward motion increases it. This motivates us to build AutoExplore Agent's reward based on this uncertainty.

\paragraph{Exploration-based training.}
We demonstrate our exploration-based training of \modelnamegt. We perform the procedure on 3 environments -  \texttt{AdventureIslandII}, which provides an easy setting for the agent to learn (single platform with no enemies at the start), \texttt{SuperMarioBros} provides an enemy and obstacles soon after the start and \texttt{Smurfs} provides a more complex background imagery and different action dynamics. For each of the environments, we train an AutoExplorer Agent. We observe that the agent learns to move forward and navigate obstacles to maximize reward. (more in Sup.Mat. D)

\begin{figure}[t]
    \centering\includegraphics[width=1\linewidth]{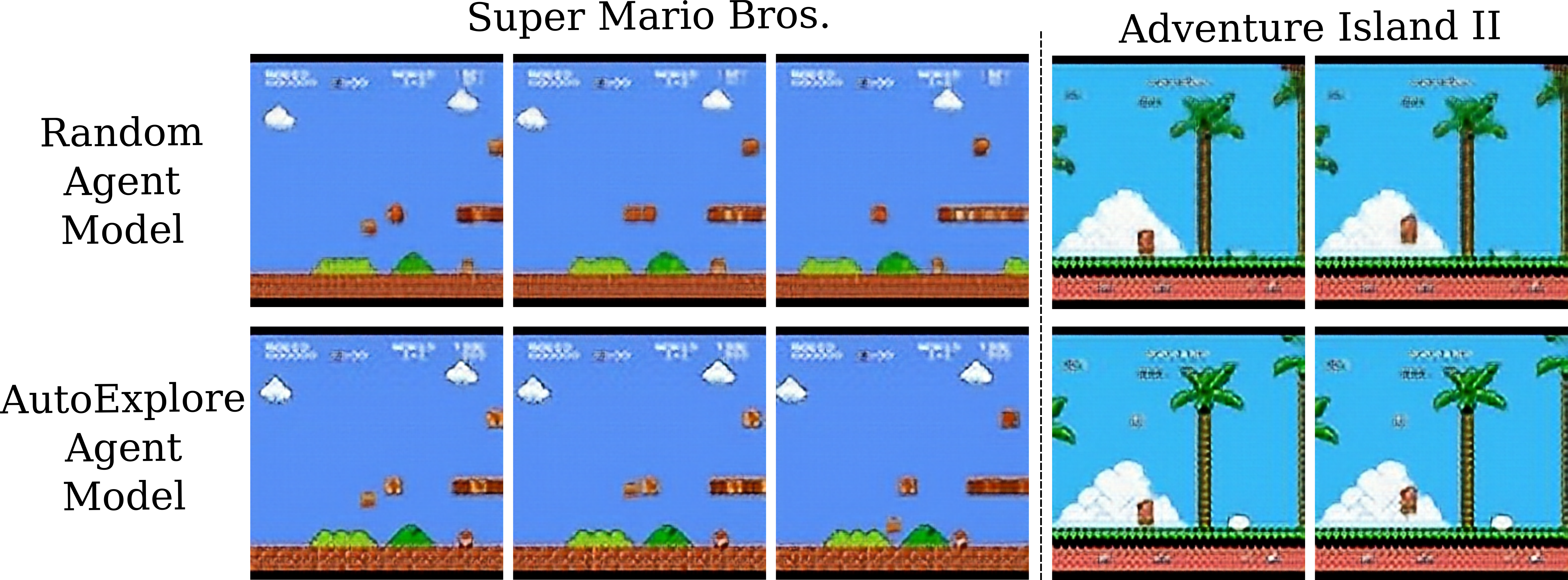}
    \vspace{-5mm}
    \caption{\textbf{AutoExplore Agent vs Random Agent Qualitative Comparison.} We show that AutoExplore exhibits better visual quality and avoids losing track of the agent.}
    \label{fig:exploration_qualitative}
    \vspace{-2mm}
\end{figure}

We use our pretrained \modelnamegt-50 model as a baseline and fine-tune it for each environment in two settings - a dataset collected on the selected environment by a random agent and by our AutoExplorer Agent. Each dataset consists of 10k sessions, each 700 frames long. We fine-tune (\modelnamegt-50-ft) for 10k iterations and pick the best performing model. In our comparison, we also include a \modelnamegt\ model trained from scratch on the diverse exploration datasets for 15k iterations to show the effect of pretraining. We perform single-pass generation for all models and the more computationally heavy autoregressive evaluation for the fine-tuned models on data from random and AutoExplore Agent's datasets.
Tab. \ref{tab:results} shows visual fidelity and controllability metrics for each environment, confirming the effectiveness of our exploration method. The model fine-tuned on AutoExplore Agent's data consistently outperforms the models trained on random actions in terms of visual fidelity. Exploration-based fine-tuning also improves controllability. Environments with small characters and uniform backgrounds can be more challenging for all models to learn. However, the gain in controllability in this case remains noticeable during autoregressive evaluation. Fig. \ref{fig:exploration_qualitative} demonstrates the superior quality of our method. In addition, we observe that the multienvironment pretraining leads to significant gains in both studied aspects compared to the nonpretrained model. (more in Sup.Mat. C)

\begin{table}[t!] 
\centering
    \centering
    \caption{\textbf{\small{Comparison of AutoExplore Agent with others.}}}
    \vspace{-3mm}
    \setlength{\tabcolsep}{1.5pt} 
    \resizebox{1.00\linewidth}{!}{
    \begin{tabular}{c|ccc|ccc}
      \hline
      \multirow{2}{*}{\textbf{Agent}} & \multicolumn{3}{c|}{\texttt{SuperMarioBros}} & \multicolumn{3}{c}{\texttt{AdventureIslandII}} \\
       & \textbf{PSNR$\downarrow$} & \textbf{SSIM$\downarrow$} & \textbf{$\Delta$PSNR$\downarrow$} & \textbf{PSNR$\downarrow$} & \textbf{SSIM$\downarrow$} & \textbf{$\Delta$PSNR$\downarrow$}\\
      \toprule
      RF & 28.58 & 0.94 & 0.181 & 24.82 & 0.78 & 0.44\\
      VAE & 24.40 & 0.86 & 0.087 & 16.57 & 0.5 & \textbf{0.072} \\ 
      Ours & \textbf{23.81} & \textbf{0.85} & \textbf{0.065} & \textbf{15.20} & \textbf{0.41} & \textbf{0.070} \\ 
      \bottomrule
    \end{tabular}
    }
    \label{tab:exploration_comparison}
    
    \vspace{-5mm}
\end{table}

\paragraph{AutoExplore Agent Evaluation.}
We compare AutoExplore Agent with exploration-based methods in \cite{ burda2018large}. We train agents based on SSE of RF and VAE features on top of \modelname\ and compare with ours on Tab. \ref{tab:exploration_comparison}.  AutoExplore Agent's reward results in maximum world model visual and controllability errors (on 1k episodes of agent actions), fulfilling its intended role in our framework.

\begin{figure}[t]
    \centering
    \begin{minipage}{0.5\textwidth}
        \centering
        \includegraphics[width=0.48\textwidth]{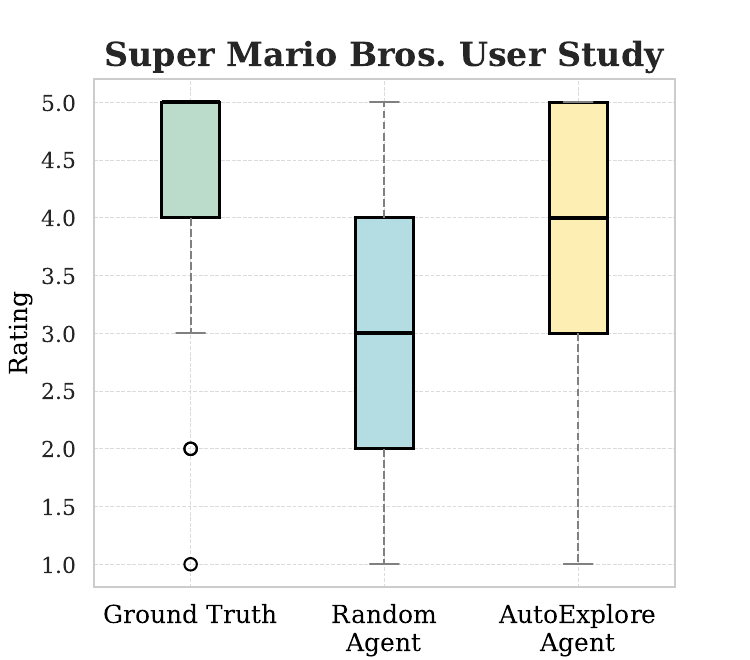}
        \includegraphics[width=0.48\textwidth]{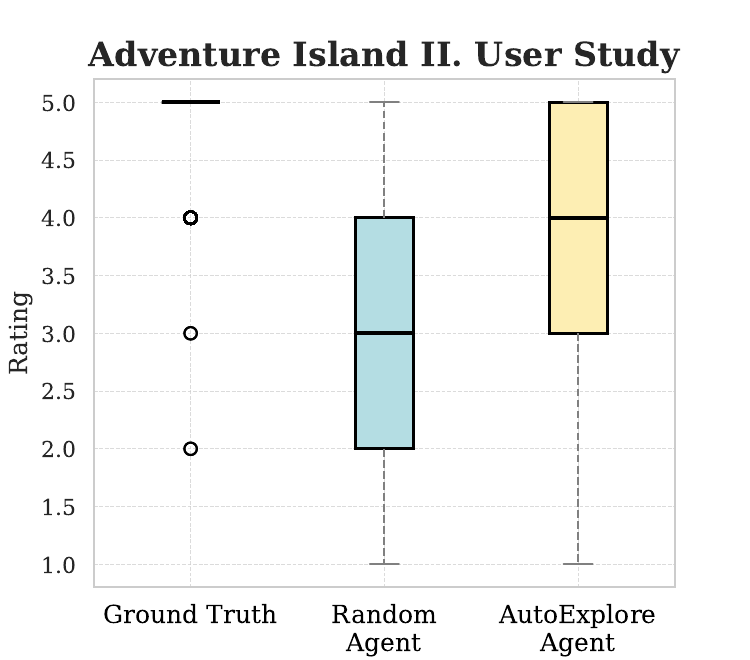}
        \vspace{-3mm}
        \caption{\textbf{User study results.} Our user study on two games shows that our model trained with AutoExplore Agent's data is consistently rated higher.}
        \label{fig:user_study}
    \end{minipage}
    \vspace{-5mm}
\end{figure}

\paragraph{User Studies.}
To validate the quality of our final results, we perform a user study in which we ask people to rate from 1 to 5 the quality of samples produced respectively by \modelnamegt\, trained on random agent's data and on AutoExplore Agent's data. Each sample in our study consists of two 16-frame clips playing in a synchronized manner - the ground truth clip and our \modelnamegt-50-ft reconstruction, given two initial frames and generating the rest autoregressively. We provide a total of 120 samples to the users - 40 samples per model and 40 samples of two ground-truth samples, to establish scale. We give 20 samples from each of the two selected games - \texttt{SuperMarioBros} and \texttt{AdventureIslandII}. We get reviews from 19 participants. The results are shown in Fig. \ref{fig:user_study}. The model, trained on data from AutoExplore Agent is clearly rated closer to the ground truth, establishing the quality of our method.

With a second user study, we evaluate the action accuracy of the generated frames. We use ambiguous single input cases (character starting mid-air) and generate 60 clips with 3 actions on \texttt{AdventureIslandII}. Users prefer our exploration-trained model, rating it \textbf{0.75 ± 0.019} on a scale from 0 (random preferred) to 1 (exploration preferred). (more in Sup.Mat. E.2)

\section{Conclusion}

As world models have developed into large models with impressive simulation properties, they require large interaction datasets, complete with diverse observations and actions. Genie \cite{bruce2024genie} demonstrates impressive abilities by training on multiple environments, however, requiring the collection of a large video dataset and a model to infer actions.

In this work, we address the heavy burden of data collection and curation by building a new framework for training large world models by collecting interaction data from a large number of virtual environments. We first build an open implementation of Genie - \modelname\ and enhance it into its version \modelnamegt. 
We obtain models exhibiting control by pretraining on a large set of virtual environments.
We address the overfitting limitations of random data collection policy by proposing AutoExplore Agent, an agent entirely independent of the environment reward, maximizing the uncertainty of \modelnamegt. After fine-tuning on the explored environment, our model is able to improve its visual fidelity and controllability much better than training solely on random agent's data. Demonstrating this on multiple environments, we show the potential of our framework to make training of next-generation world models more accessible, cost-effective, and effort-free.

\section{Acknowledgments}
INSAIT, Sofia University "St. Kliment Ohridski". Partially funded by the Ministry of Education and Science of Bulgaria’s support for INSAIT as part of the Bulgarian National Roadmap for Research Infrastructure. This project was supported with computational resources provided by Google Cloud Platform (GCP).

{
    \small
    \bibliographystyle{ieeenat_fullname}
    \bibliography{main}
}


\maketitlesupplementary
\appendix

\setcounter{page}{1} 
\renewcommand{\thepage}{S\arabic{page}}

\setcounter{section}{0}

\part{} 
\parttoc 



\section{Experimental Protocol}

\subsection{Training Protocol  of \modelname\ and \modelnamegt.}\label{sec:training_gt}

The architecture and training parameters of the Tokenizer and the Dynamics module of \modelnamegt\ are shown respectively in the Tab. \ref{tab:tokenizer_hpparams} and Tab. \ref{tab:dynamics_hpparams}. \modelname\ shares those choices, with the addition of LAM defined as in Tab. \ref{tab:lam_hpparams}. For the purpose of the case study, we use 7 latent actions. Training parameters can be seen in Tab. \ref{tab:opt_hpparams}.

\begin{table}[t!]
    \centering
    \begin{tabular}{ccc}
    \hline
    \textbf{Component} & \textbf{Parameter} & \textbf{Value} \\ \hline
    Encoder            & num\_blocks        & 8              \\
                       & d\_model           & 512            \\
                       & num\_heads         & 8              \\ \hline
    Decoder            & num\_block        & 8              \\
                       & d\_model           & 512            \\
                       & num\_heads         & 8              \\ \hline
    Codebook           & num\_codes         & 1024           \\
                       & latent\_dim        & 32             \\ \hline\\
    \end{tabular}
    \vspace{-5mm}
    \caption{Tokenizer hyperparameters}
    \label{tab:tokenizer_hpparams}
\end{table}

\begin{table}[t!]
    \centering
    \begin{tabular}{ccc}
        \hline
    \textbf{Component} & \textbf{Parameter} & \textbf{Value} \\ \hline
    Architecture            & num\_blocks        & 12              \\
                       & d\_model           & 512            \\
                       & num\_heads         & 8              \\
    Sampling            & temperature       & 1.0              \\
                       & maskgit\_steps         & 25              \\ \hline\\
    \end{tabular}
    \vspace{-5mm}
    \caption{Dynamics hyperparameters}
    \label{tab:dynamics_hpparams}
\end{table}

\begin{table}[t!]
    \centering
    \begin{tabular}{ccc}
        \hline
    \textbf{Component} & \textbf{Parameter} & \textbf{Value} \\ \hline
    Encoder            & num\_blocks        & 8              \\
                       & d\_model           & 512            \\
                       & num\_heads         & 8              \\ \hline
    Decoder            & num\_blocks        & 8              \\
                       & d\_model           & 512            \\
                       & num\_heads         & 8              \\ \hline
    Codebook           & num\_codes         & 7           \\
                       & latent\_dim        & 32             \\ \hline\\
    \end{tabular}
    \vspace{-5mm}
    \caption{LAM hyperparameters}
    \label{tab:lam_hpparams}
\end{table}

\begin{table}[t!]
\centering
\begin{tabular}{cc}
\hline
\textbf{Parameter}                  & \textbf{Value}  \\ \hline
max\_lr    & $1 \times 10^{-4}$ \\
min\_lr     & $5 \times 10^{-5}$ \\
$\beta_1$   & 0.9              \\
$\beta_2$  & 0.99             \\
weight\_decay                        & $1 \times 10^{-4}$ \\
linear\_warmup\_start\_factor          & 0.5              \\
warmup\_steps                        & 5000            \\ \hline\\
\end{tabular}
    \vspace{-5mm}
\caption{Optimizer Hyperparameters}
\label{tab:opt_hpparams}
\end{table}

We train the Tokenizer on 8 A100 GPUs for 72k iterations, with batch size 112 and patch size 4, on a dataset of all 483 environments (50 sessions per environment obtained with a random agent). 

Our Dynamics module is trained with sequences of 16 frames, processed by the pretrained tokenizer. Dynamics module is trained with batch size 80 on 8 A100 GPUs for 185k iterations on \texttt{Platformers-200} (\modelnamegt-200), and fine-tuned for 80k iterations on \texttt{Platformers-50} (\modelnamegt-50), batch size 160. 

For an agent (random or AutoExplore Agent), we obtain a dataset of 10k sequences of length 800. We fine-tune \modelnamegt-50 on a set for 10k iterations to obtain \modelnamegt-50-ft, with batch size 160.

We always use the Adam optimizer with a linear warm-up and cosine annealing strategy.

We note that GenieRedux has \(\sim\) 350M total parameters, broken down as follows: Tokenizer (100M), LAM (170M), and Dynamics (80M). Meanwhile, GenieRedux-G has \(\sim\)180M total parameters: Tokenizer (100M) and Dynamics (80M).

\subsection{Testing Protocol  of \modelname\ and \modelnamegt.}

For our test set, we train Agent57 per environment, using the available environment reward. In order to have many diverse episodes in our datasets and all the actions to be represented, we mix, using an $\epsilon$-greedy approach, the agent's actions with random actions.
We collect 1000 episodes as the test set (with episode length 700) and evaluate on sequences of size 12 with step size 20 in two settings. While our model can handle a single frame as input, for a fair evaluation, we choose to provide two, as a single frame does not provide motion information and there are multiple valid solutions (see Sect. \ref{sec:limitations}). We provide two frames and predict the next 10, given all actions. In the usual case, we perform MaskGIT inference with 25 iterations for all 10 images at once. We obtain much fewer artifacts and higher level of control if we adapt an autoregressive approach - iteratively generating 2 frames at a time given all previous tokens in the sequence, each with 25 iterations. However, as this is computationally heavy, we provide autoregressive results for our best models only in our evaluations.

\subsection{Training Protocol of AutoExplore Agent}

For each of the environments, we train for 300 epochs with the following schedule for each epoch: 1. Run the current agent for 200 steps in 8 running environments in parallel to collect data in the replay buffer. Actions are sampled with temperature 1.0, with an epsilon-greedy algorithm with $\epsilon$ starting from 0.1 and linearly decaying to 0.01 over the course of 150 epochs. 2. Train the agent for 200 steps, sampling from the buffer, with batch size 128. Actor-critic loss is used with entropy regularization over the actions to prevent greedy behavior. In the end, we choose the agent with the highest evaluation return throughout training.

\section{Super-Resolution Network}
We upscale the outputs of \modelnamegt\ from 64x64 to 256x256 by a U-Net based super-resolution network, with MSE loss for both training and evaluation. The training data consists of 256x256 images that we captured from the original environments. Three configurations were tested: (1) a small U-Net with feature channel dimensions [64,128,256,512] and approximately 31 million parameters, trained on 16,000 images (12,000 for training and 4,000 for testing), achieving a test loss of 0.0081; (2) the same small U-Net trained on a larger dataset of 50,000 images (45,000 for training and 5,000 for testing), achieving a test loss of 0.0047; and (3) a larger U-Net with feature channel dimensions [128,256,512,1024] and 124 million parameters, trained on the same 50,000-image dataset, achieving a test loss of 0.0029. All models were trained with a batch size of 128, a learning rate of 0.0001, and a step-based scheduler (step\_size=25, gamma=0.5) for 300 epochs, using Adam optimizer.

\begin{figure*}[!t]
   \includegraphics[width=1\textwidth]{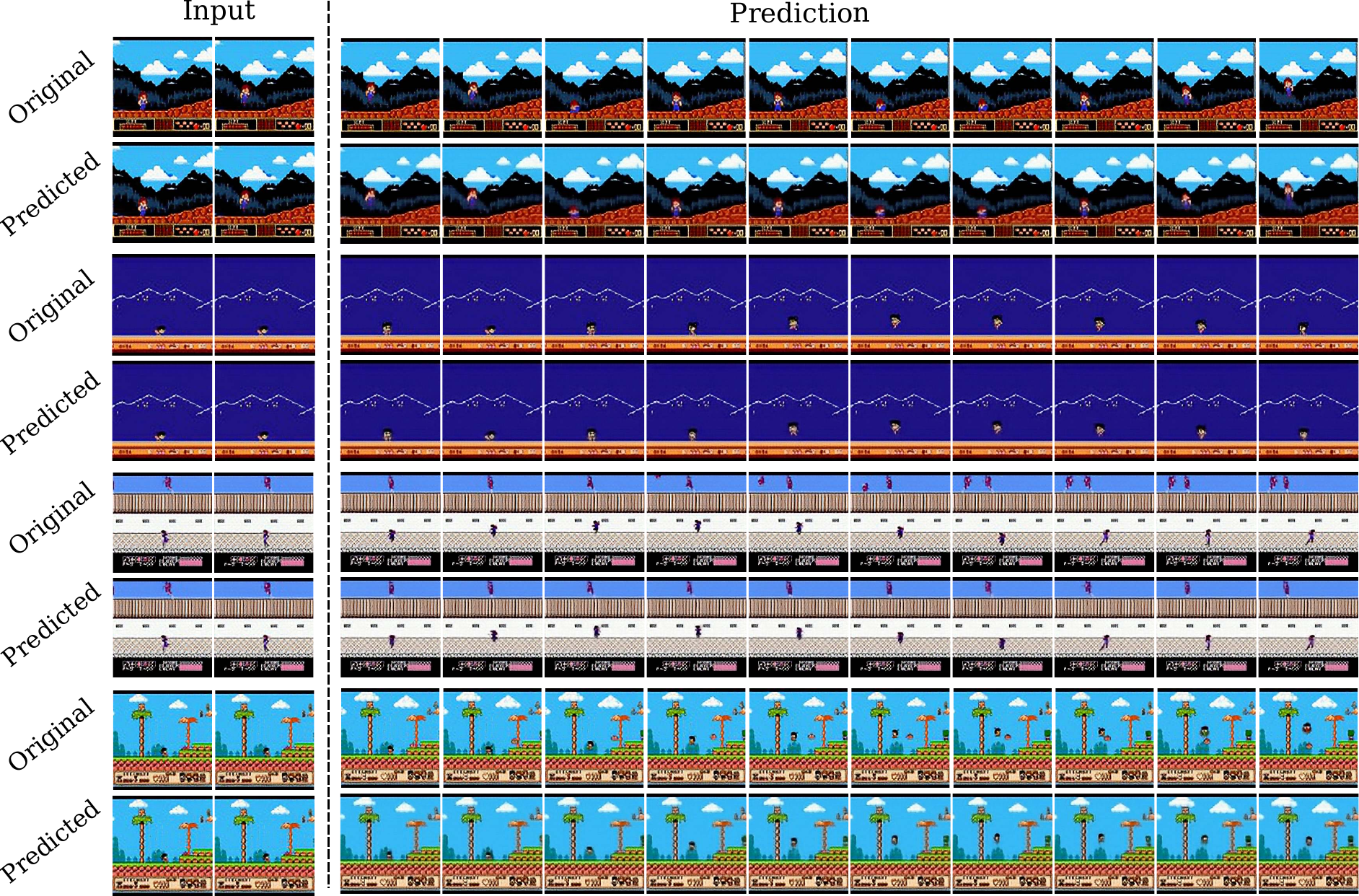}
   \caption{\textbf{Qualitative Results of \modelnamegt-50.} Given 2 frames, we show the predictions of 10 future frames. Ground truth is shown for comparison.}
   \label{fig:multienv_50_repeat}
\end{figure*}

\section{Multi-Environment Models Additional Experiments}

\subsection{Qualitative Results of \modelnamegt-50}

\begin{figure*}[t]
   \includegraphics[width=1\textwidth]{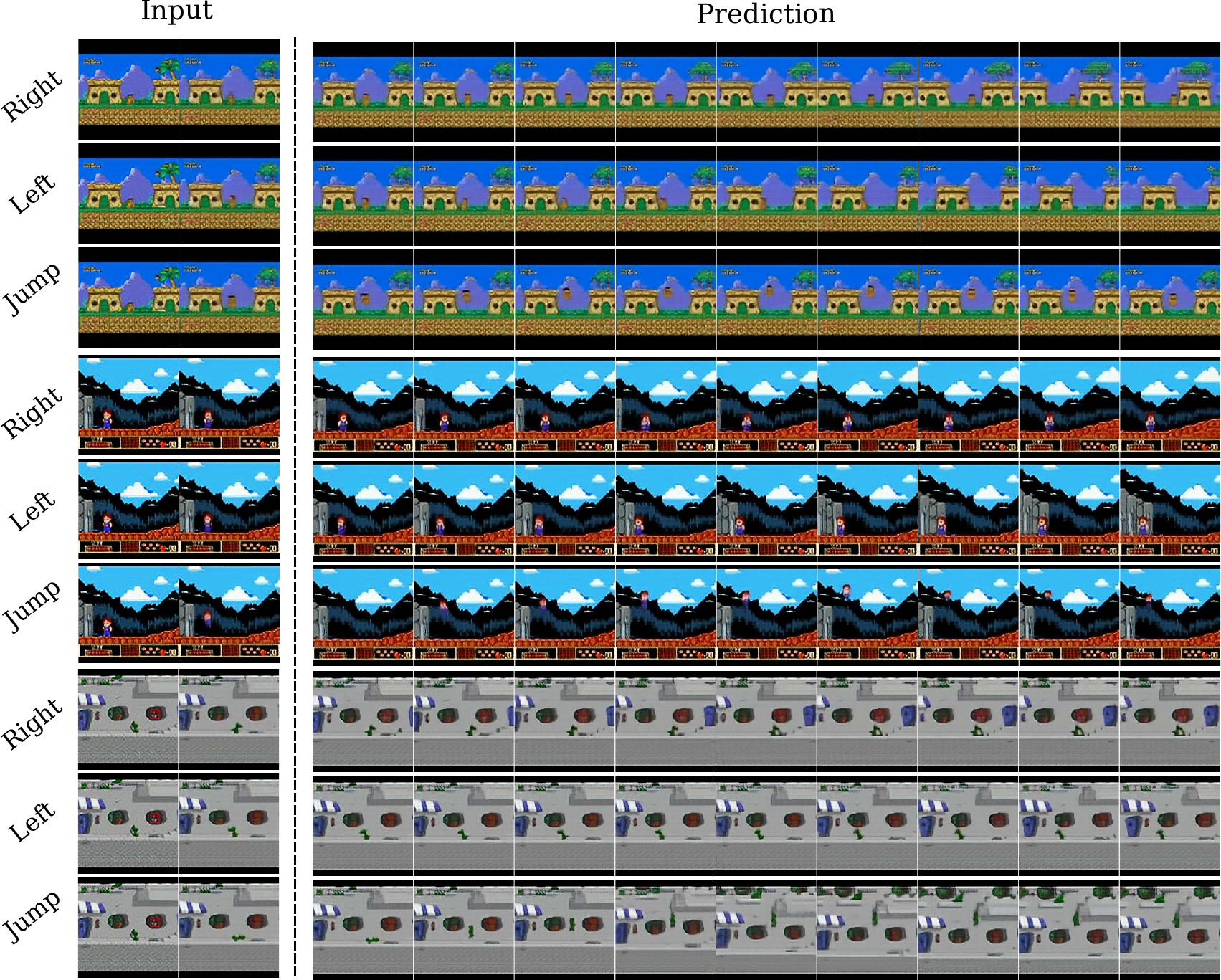}
   \caption{\textbf{Controllability of \modelnamegt-50.} We show a detailed per frame prediction for right, left and jump actions, showing the development of the actions. }
   \label{fig:multienv_50_control}
\end{figure*}

In Fig. \ref{fig:multienv_50_repeat} we show examples of 10-frame predictions from \modelnamegt-50. They are sampled from the test set and the actions that resulted in the ground truth sequence were given to the model to produce the predictions. As seen, the model was able to produce outcomes from the actions that are close to the ground truth. In Fig. \ref{fig:multienv_50_control} is shown the developments of 3 actions over time for \modelnamegt-50, showing a smooth trajectory and the action being executed. 

In our experiments, we test the ability of our models to simulate multiple environments in virtual environments already observed by the models. For new unseen environments, our models show limited generalization abilities, characterized by pausing motion and visual artifacts. We believe that generalizability can be improved by training our tokenizer on a larger video dataset, however, with care taken to preserve the learned background token strategy learned, as it brings important properties to our exploration reward and the Dynamics module.

\begin{table*}[!t]
    \centering
    \caption{\textbf{\texttt{SuperMarioBros} Autoregressive Quantitative Evaluation.}}
    \vspace{-2mm}
    \begin{tabular}{c|c|c|cccc}
        \hline
        \textbf{Environment} & \textbf{Strategy} & \textbf{Model} & \textbf{FID$\downarrow$} & \textbf{PSNR$\uparrow$} & \textbf{SSIM$\uparrow$} & \textbf{$\Delta$PSNR$\uparrow$} \\
        \hline
        \multirow{5}{*}{Super Mario Bros.} & \multirow{2}{*}{Random Autoregressive} & \modelnamegt-50  & 30.48 & 34.59 & 0.94 & 0.55 \\
        &  & \modelnamegt-50-ft  & 30.84 & 34.85 & 0.95 & 0.57 \\
         \cline{2-7}
         & \multirow{3}{*}{Exploration Autoregressive}  &
         Tokenizer-ft & 8.08 & 42.00 & 0.99 & - \\
        & &\modelnamegt\ & 9.46 & 34.38 & 0.95 & 0.07 \\ 
        & & \modelnamegt-50-ft & \textbf{9.33} & \textbf{37.77} & \textbf{0.97} & \textbf{0.76} \\
        
        \hline
    \end{tabular}
    \label{tab:mario_quantitative}
\end{table*}

\begin{table*}[!t]
    \centering
    \caption{\textbf{Quantitative results of \modelnamegt-50, fine-tuned on AutoExplore Agent's data of all three environments together.} Results show that our method can be used to improve the multi-environment performance of the model. \modelnamegt-50 is our pretrained world model on 50 environments. \modelnamegt-50-ft are fine-tuned models using data from a random agent or AutoExplore (Exploration). \modelnamegt\ denotes a non-fine-tuned model, trained with the exploration data. }
    \vspace{-2mm}
    \begin{tabular}{c|c|c|cccc}
        \hline
        \textbf{Environment} & \textbf{Strategy} & \textbf{Model} & \textbf{FID$\downarrow$} & \textbf{PSNR$\uparrow$} & \textbf{SSIM$\uparrow$} & \textbf{$\Delta$PSNR$\uparrow$} \\
        \hline
        & \multirow{2}{*}{Random} & \modelnamegt-50  & 43.57 & 27.55 & 0.82 & 0.65 \\
        & & \modelnamegt-50-ft & 43.98 & 27.74 & 0.82 & 0.78 \\
         \cline{2-7}
        \multirow{3}{*}{Combined Environments} & \multirow{3}{*}{Exploration}  &
         Tokenizer-ft & 14.02 & 37.98 & 0.98 & - \\
        & &\modelnamegt\ & 14.88 & 28.91 & 0.88 & 0.25 \\ 
        & & \modelnamegt-50-ft & \textbf{14.61} & \textbf{31.29} & \textbf{0.91} & \textbf{1.09} \\
        
         \cline{2-7}
        & Random Autoregressive & \modelnamegt-50-ft  & 43.69 & 28.19 & 0.83 & 0.79 \\
        & Exploration Autoregressive& \modelnamegt-50-ft  & \textbf{14.49} & \textbf{33.14} & \textbf{0.93} & \textbf{1.46} \\
        \hline
    \end{tabular}
    \label{tab:multienv_quantitative}
\end{table*}

\subsection{Autoregressive Evaluation}

For one of the environments - \texttt{SuperMarioBros}, we provide comparison of all our models using autoregressive evaluation. This evaluation is more computationally heavy, so we originally compare them with a single-pass evaluation, and evaluate autoregressively only the fine-tuned models on both strategies - with a random agent and AutoExplore Agent. As for small characters and uniform backgrounds, single-pass evaluation appears to produce close results between all models in terms of controllability, we choose to perform full autoregressive evaluation on \texttt{SuperMarioBros} (where these conditions are present) to show the benefit of our approach. Results are shown in Tab. \ref{tab:mario_quantitative}. The newly autoregressively evaluated models are \modelnamegt-50 and \modelname. It can be concluded that, with our exploration approach, we obtain significantly better results in terms of visual fidelity and controllability.

\subsection{Multi-Environment Fine-tuning}

In this experiment, we take the diverse datasets, collected from the three environments we have studied - \texttt{AdventureIslandII}, \texttt{SuperMarioBros} and \texttt{Smurfs}, and fine-tune \modelnamegt-50 on all of them together. In this way, we evaluate the effect of our method on multi-environment training. Results are shown in Tab. \ref{tab:multienv_quantitative}.  Using AutoExplore Agent's data, the model has improved its visual fidelity and controllability across the test set, containing all three environments (equal number of samples each). This shows that our method is applicable for improving multi-environment training as well.

\begin{figure*}[!t]
   \includegraphics[width=1\textwidth]{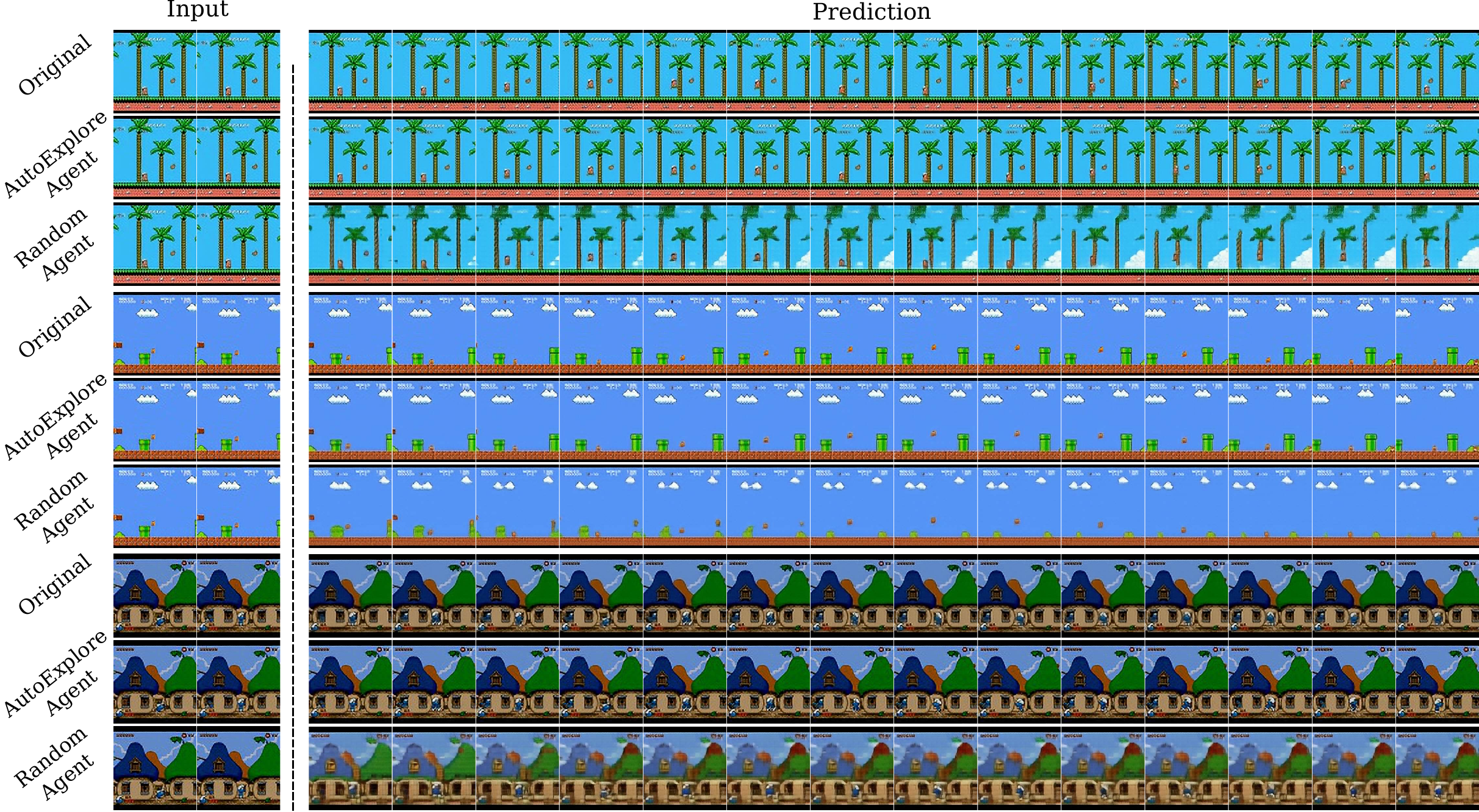}
   \caption{\textbf{Qualitative Results of \modelnamegt-50-ft for random and AutoExplore agents.} They are compared to the ground truth (Original). It can be seen that AutoExplore Agent's data produces a model with significantly higher visual quality and less artifacts. }
   \label{fig:ft_qualitative}
\end{figure*}

\begin{figure*}[!t]
   \includegraphics[width=1\textwidth]{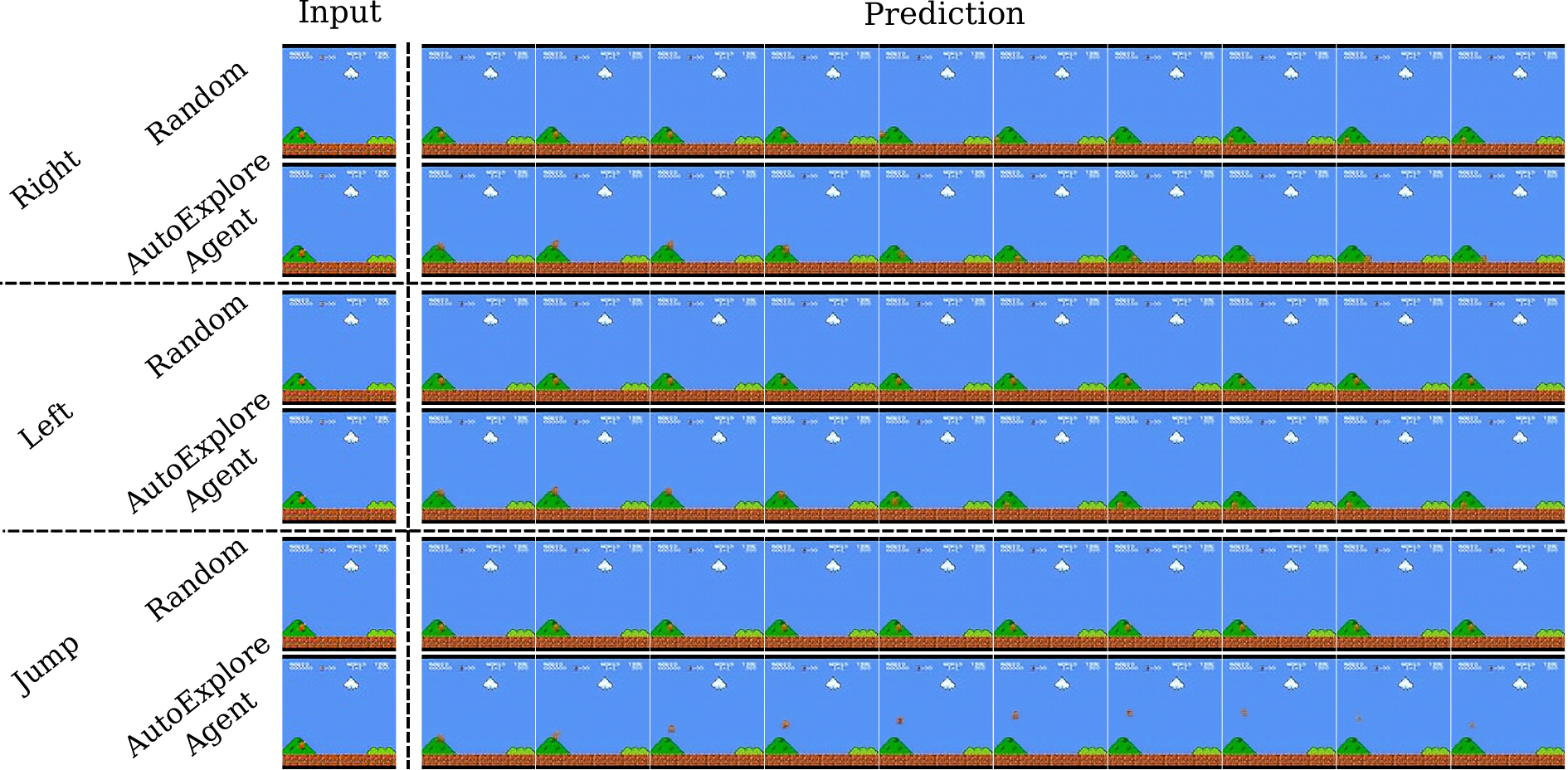}
   \caption{\textbf{Controllability of \modelnamegt-50-ft for random and AutoExplore agents on \texttt{SuperMarioBros}.} In ambiguous cases like this where the agent can be going up or down, exploration data shows to improve performance. }
   \vspace{-5mm}
   \label{fig:ft_control}
\end{figure*}

\subsection{Qualitative Evaluation of Fine-tuned Models}

In Fig. \ref{fig:ft_qualitative} are shown examples per environment of predictions from a model, fine-tuned on a random agent versus a model fine-tuned on AutoExplore Agent's data. The model, trained on AutoExplore Agent's data, exhibits much higher visual quality and less artifacts.
We also note that the tokenizer plays a role in improving visual quality. After exploration, the tokenizer is able to fit better to new visuals of the environment, which reduces visual artifacts.

In Fig. \ref{fig:ft_control} we show AutoExplore Agent's data helping to achieve better controllability compared to the random agent. As typical for controllability evaluation, we give a single frame as input. We observe that in cases where the motion is ambiguous (e.g. where a character might be going up or down), fine-tuning with exploration data leads to more confidence and hence realistic sequence generation. In contrast, models trained on random data cannot resolve the situation and copy the frame multiple times.

\section{AutoExplore Agent Behavior Study}
\label{sec:agent_bahvior_study}

\begin{figure}[t!]
    \centering
    \includegraphics[width=1\linewidth]{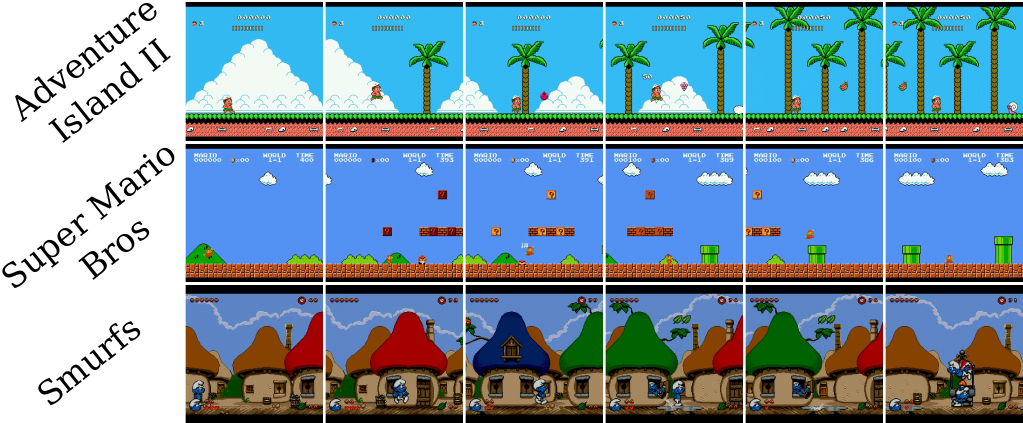}
    \caption{\textbf{AutoExplore Agent Behavior.} We show the behavior of our AutoExplore Agent on the three environments studied. It can be seen that the agent learned to progress by moving right, jumping over obstacles and dealing with enemies.}
    \label{fig:agent_behavior}
\end{figure}

The agent was trained with the five actions that the world model was trained with. While initially in training the agent learns simpler strategies like jumping, eventually it achieves better returns by learning to move forwards in an environment (and reveal new scenes). To progress even further, the agent learns to overcome obstacles and enemies. In Fig. \ref{fig:agent_behavior}, we show the behavior of the agent on the three environments used after training. The agent is observed to move forward in the environment, to overcome enemies, jump over obstacles. Interestingly, the strategy in Smurfs was to sometimes wait to be attacked by an enemy, which caused the player to disappear and the camera to move before spawning. This seems to cause an increase in world model uncertainty in that environment. In other cases (flying enemies), the agent tries to avoid. In Smurfs, there is an action of entering a door. We observe that sometimes the agent enters a door that causes the character to reappear from a different side on the screen.

\section{User Studies}
\subsection{General Quality User Study Details}

\begin{figure}[t!]
    \centering
    \includegraphics[width=1\linewidth]{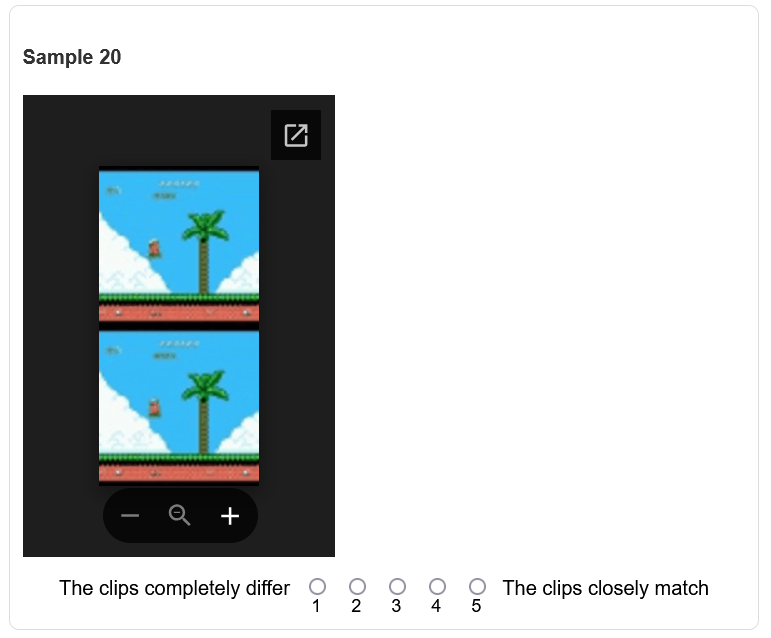}
    \caption{\textbf{General User Study Sample.}}
    \label{fig:user_study_extra}
\end{figure}

We provide extra details about our user study to evaluate the models fine-tuned on data from a random agent and from AutoExplore Agent. In Fig. \ref{fig:user_study_extra} we show the interface for a single sample given to the user. A clip is shown with two parts that the user should compare and rate. The order of the samples is random. The instructions given to the users at the start of the study are provided below.

\textit{
Thank you for participating in our study! You will watch a total of 120 video samples. Each sample consists of two clips:\\
\textbf{Top} clip: Reference\\
\textbf{Bottom} clip: Comparison clip\\
Please compare the two clips in each sample and rate how closely they match in terms of visual quality and content.\\
Use the scale provided:\\
\textbf{1} : The two clips completely differ in terms of visual quality and/or content\\
\textbf{5} : The two clips closely match in terms of visual quality and content\\
Submit your rating for each sample through this form. Your feedback is important and greatly appreciated!
}

\begin{figure}[t!]
    \centering
    \includegraphics[width=1\linewidth]{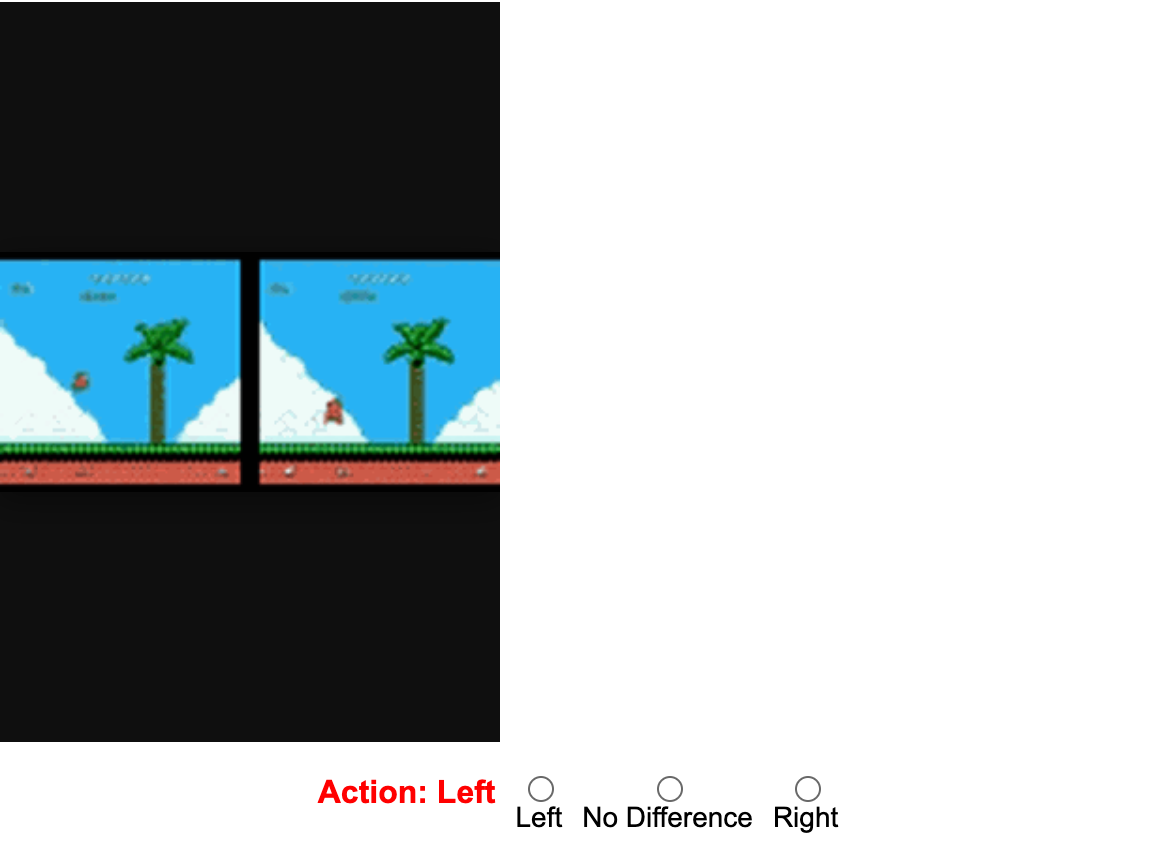}
    \caption{\textbf{Action Quality User Study Sample.}}
    \label{fig:action_user_study_extra}
\end{figure}

\subsection{Action Quality User Study Details}

We conduct a second user study to specifically evaluate the gains in action quality of our model fine-tuned on data from the AutoExplore agent over the baseline. Observing that our model is particularly beneficial in scenarios with ambiguous initial frames, we use this user study to test this. We use single initial frames with the agent in mid-jump. Participants interact with an interface shown in Fig.\ref{fig:action_user_study_extra}. The user is shown pairs of synchronized videos, generated by the baseline and the exploration model (left/right position is randomized). Both videos depicted the same action — \emph{left}, \emph{jump}, or \emph{right}—which was explicitly labeled in \emph{bold red} below them. Participants were instructed to assess the quality of the action performed, disregarding any differences related purely to visual quality, and select one of the following options:

\begin{itemize}
    \item \emph{\textbf{Left:} The left clip depicts the action more accurately.}
    \item \emph{\textbf{No Difference:} Both clips depict the action equally well.}
    \item \emph{\textbf{Right:} The right clip depicts the action more accurately.}
\end{itemize}

\section{\modelname\ Evaluation on CoinRun Case Study}

In this section, we qualitatively evaluate our Genie implementation - \modelname\, and its variant \modelnamegt\ on the CoinRun case study. We also quantitatively and qualitatively study the effect of using data from a trained agent in the Coinrun environment (\modelname\ and \modelnamegt). We study the behavior and limitations of the model and compare our implementation with a concurrent one.

\subsection{CoinRun Case Study Details}

We train the Tokenizer and the Dynamics module on CoinRun environment datasets, one obtained from a random agent, and one obtained from a trained agent using environment reward.

For training the agent for exploration, we enable velocity maps on CoinRun. These maps also need to be enabled for the agent during data collection. When evaluating models trained on different datasets (random agent vs. trained agent), to be fair, we exclude the velocity map regions by setting their pixels to black on all sets. 

Throughout the training, we use a batch size of 84 and a patch size of 4 for all components. We use the Adam Optimizer with a linear warm-up and cosine annealing strategy.

We refer to the test set obtained from a random agent as \randomtestname\ and to the one obtained from a trained agent as \ppotestname.

\begin{table}
    \centering
    \captionsetup{justification=centering}
    \caption{\textbf{Visual Fidelity} of TA models.}
    \begin{tabular}{c|ccc}
      \hline
      \multirow{2}{*}{\textbf{Model}} & \multicolumn{3}{c}{\textbf{\randomtestname}} \\
       & \textbf{FID$\downarrow$} & \textbf{PSNR$\uparrow$} & \textbf{SSIM$\uparrow$} \\
      \toprule
      Tokenizer-TA & 12.10 & 39.53 & 0.97 \\
      LAM-TA & 47.73 & 28.24 & 0.85 \\
      \midrule
      \modelname-TA & 13.26 & 25.47 & 0.82 \\
      \modelname-G-TA & \textbf{13.01} & \textbf{32.09} & \textbf{0.94} \\
      \bottomrule
    \end{tabular}
    \label{tab:ta_fidelity_eval}
\end{table}


\subsection{Prediction Horizon Evaluations}

\begin{figure}
    \centering
    \caption{\textbf{\modelnamegt-TA  Controllability Across Horizons.}}
    \includegraphics[width=0.5\linewidth]{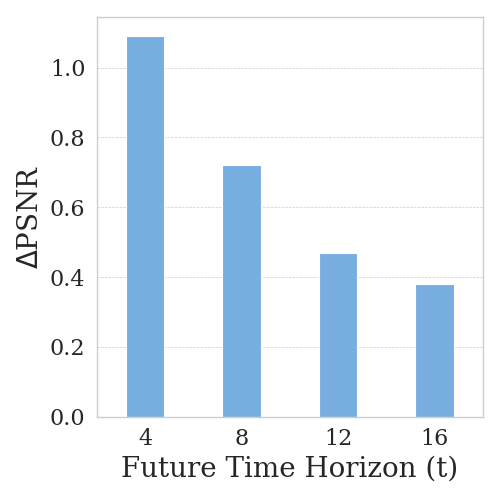}
    \label{fig:horizons}
\end{figure}

We evaluate the controllability of our best model (at 50k iterations) over varying prediction horizons in Fig. \ref{fig:horizons}. As expected, predictions become more challenging further into the future. The first prediction is also difficult due to insufficient motion information - we obtain 0.4 $\Delta_t$PSNR for $t=1$. To address this issue, we provide the model with 4 frames and actions (predicting 10), and observe an improvement of our best model (\modelname-G-TA) from 34.79 PSNR (12.75 FID) in our results in the main paper to \textbf{38.31} PSNR (12.29 FID) on \ppotestname.

\subsection{\modelnamegt\ Qualitative Evaluation}\label{sec:app_genieredux_base_examples}

\begin{figure}[t!]
    \centering
    \includegraphics[width=1\linewidth]{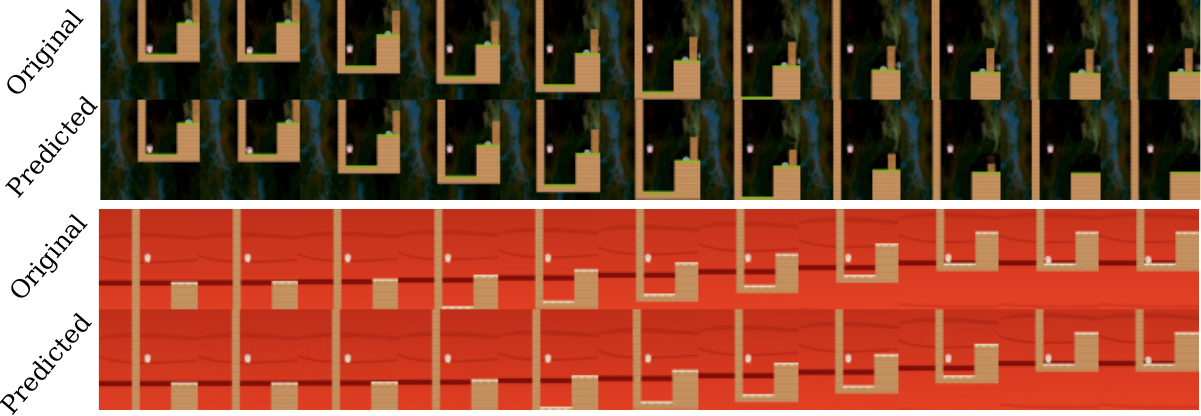}
    \caption{\textbf{\modelname\ Quantitative Evaluation.} We present a few sequences from the test set with predictions from \modelname. On the example at the top we show a successful jump action. On the example at the bottom we show a successful motion progression.}
    \label{fig:genieredux_base_examples}
\end{figure}

In Fig. \ref{fig:genieredux_base_examples} we show quantitative results demonstrating that \modelnamegt\ can perform motion progression and action execution.

\subsection{\modelname-TA Qualitative Evaluation}\label{sec:app_genie_jump_demo}

\begin{figure}[h!]
    \centering
    \includegraphics[width=1\linewidth]{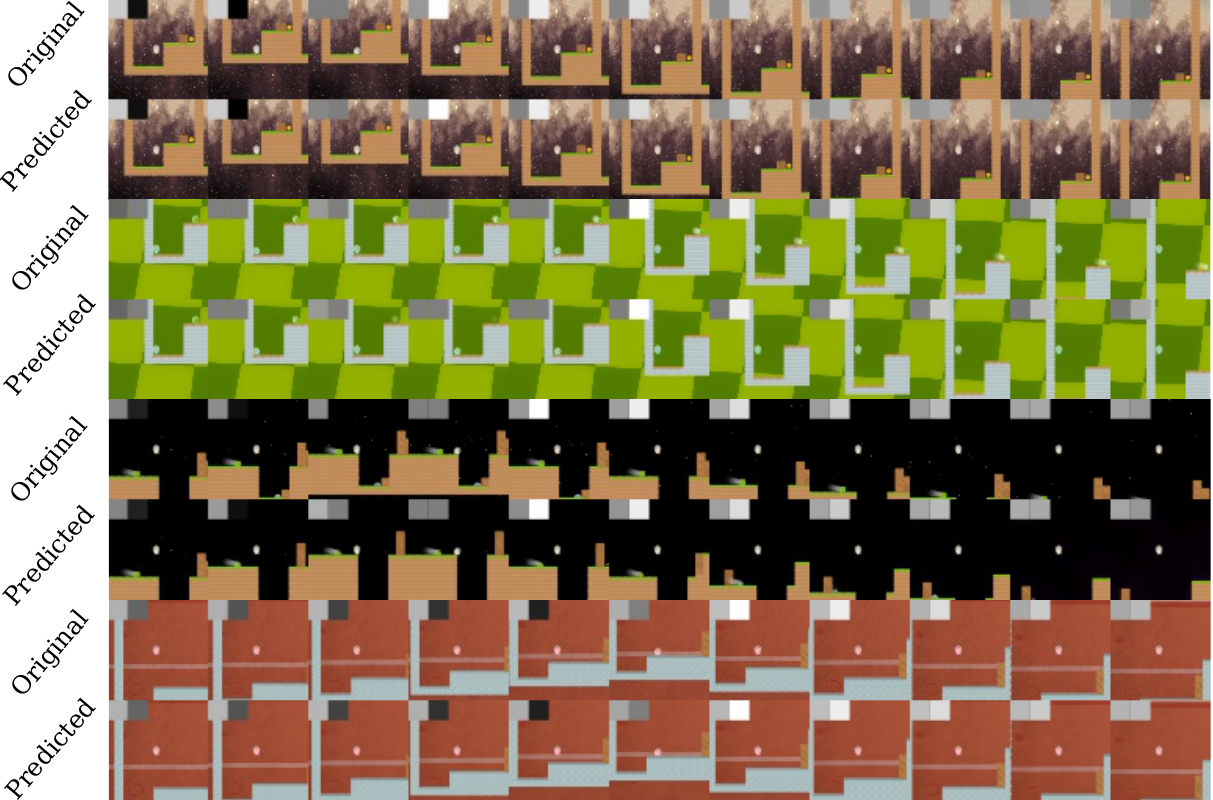}
    \caption{\textbf{GenieRedux-TA Qualitative Comparison.} We present a few samples from the test set with various actions. We demonstrate that \modelname-TA performs the actions correctly.}
    \label{fig:genieredux_ta_jump_demo}
\end{figure}

\begin{figure}[h!]
    \centering
    \includegraphics[width=1\linewidth]{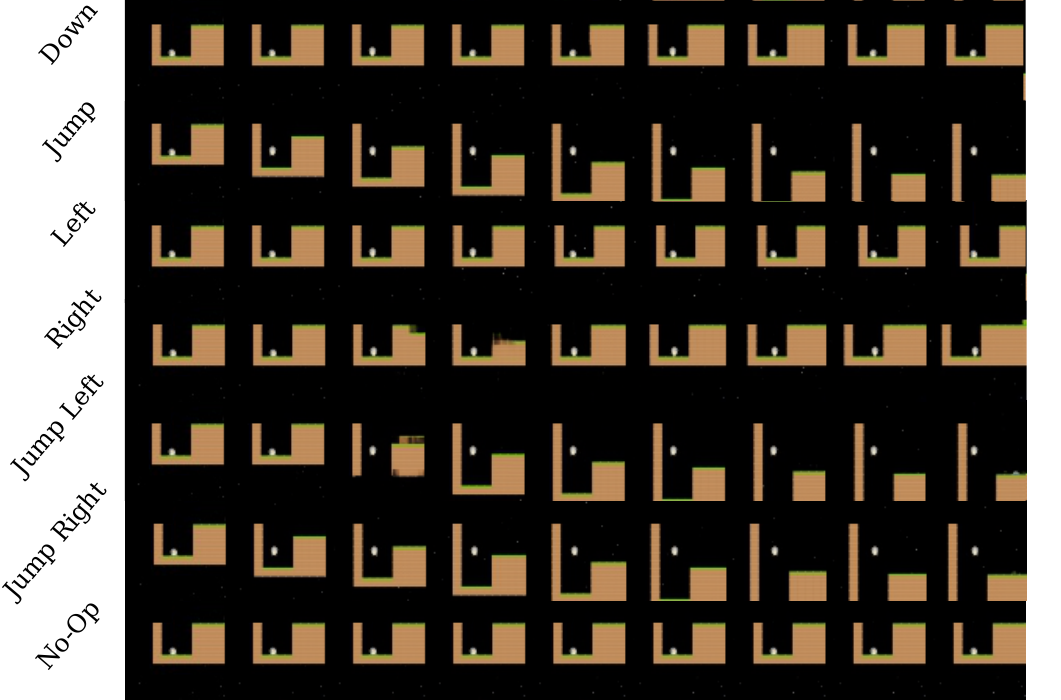}
    \caption{\textbf{GenieRedux-TA Controllability.} We show predictions for all environment actions of \modelname-TA.}
    \label{fig:genieredux_ta_controllability}
\end{figure}

In Fig. \ref{fig:genieredux_ta_jump_demo} we demonstrate that \modelname-TA is able to execute actions and complete motion. In Fig. \ref{fig:genieredux_ta_controllability} we show that the model is capable of executing all actions of the environment.


\subsection{Jafar Qualitative Comparison}\label{sec:jafar_qualitative_results}
We compare with Jafar \citesup{willi2024jafar} - a concurrent with our implementation of Genie (in JAX). We obtain and train their model as instructed. We train \modelname\ with Jafar's model parameters and like them separate LAM from Dynamics in training. The latter significantly worsened \modelname's action representation. Despite that, \modelname\ shows
significantly better visual fidelity metrics, achieving 17.91 PSNR (46.12 FID), compared to Jafar's 12.66 PSNR (154.12 FID). \modelname\ does not exhibit Jafar's artifacts or the reported problematic "hole digging" behavior. Moreover, we observe that Jafar lacks causality, which we find problematic.

\begin{figure}[t!]
    \centering
    \includegraphics[width=1\linewidth]{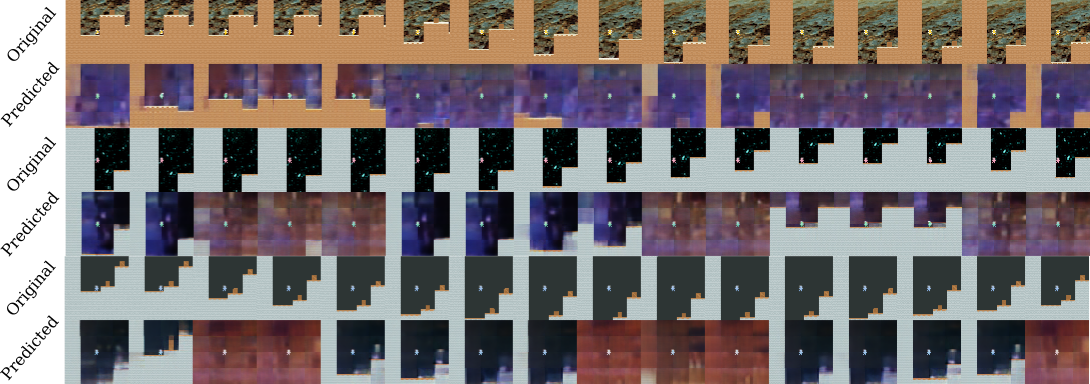}
    \caption{\textbf{Jafar Qualitative Results.} The results are on the validation set. We give only a single image and actions and predict 15 frames in the future.}
    \label{fig:jafar_qualitative_results}
\end{figure}

In Fig. \ref{fig:jafar_qualitative_results} we show Jafar's reconstruction of 10 frames into the future, given the first frame and a sequence of actions. The results are on the validation set after training. We observe an abundance of artifacts.
We note that if we provide the images instead of providing the first frame, we get much less artifacts. This seems to hint that Jafar relies on future images to make predictions for the current frame, which might be an inherent problem of the model not being causal.

We additionally report test set results for Jafar - 0.48 SSIM and for \modelname\ (with Jafar parameters) - 0.62 SSIM.

\begin{figure}[t!]
    \centering
    \includegraphics[width=1\linewidth]{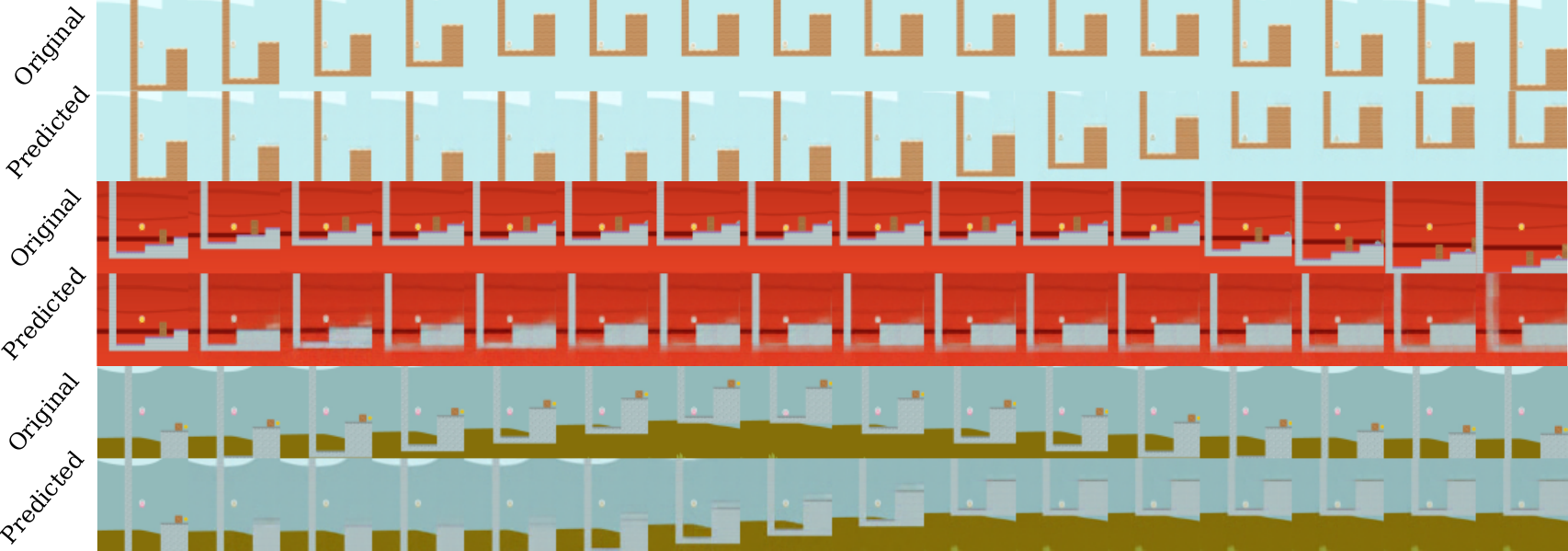}
    \caption{\textbf{\modelname\ with Jafar's Parameters Qualitative Results.} We show 15 frames into the future given actions and an initial frame of our model.}
    \label{fig:jafar_our_qualitative_results}
\end{figure}

In addition, we show the version of \modelname\ that we trained to match Jafar in Fig. \ref{fig:jafar_our_qualitative_results}. While it can be noticed that the model prefers inaction when encountering actions, it successfully progresses motion - e.g. moving a character through the air. We also notice fairly good visual quality.

\subsection{Additional \modelnamegt-TA Qualitative Results and Limitations}\label{sec:limitations}

\begin{figure}[t!]
    \centering
    \includegraphics[width=1\linewidth]{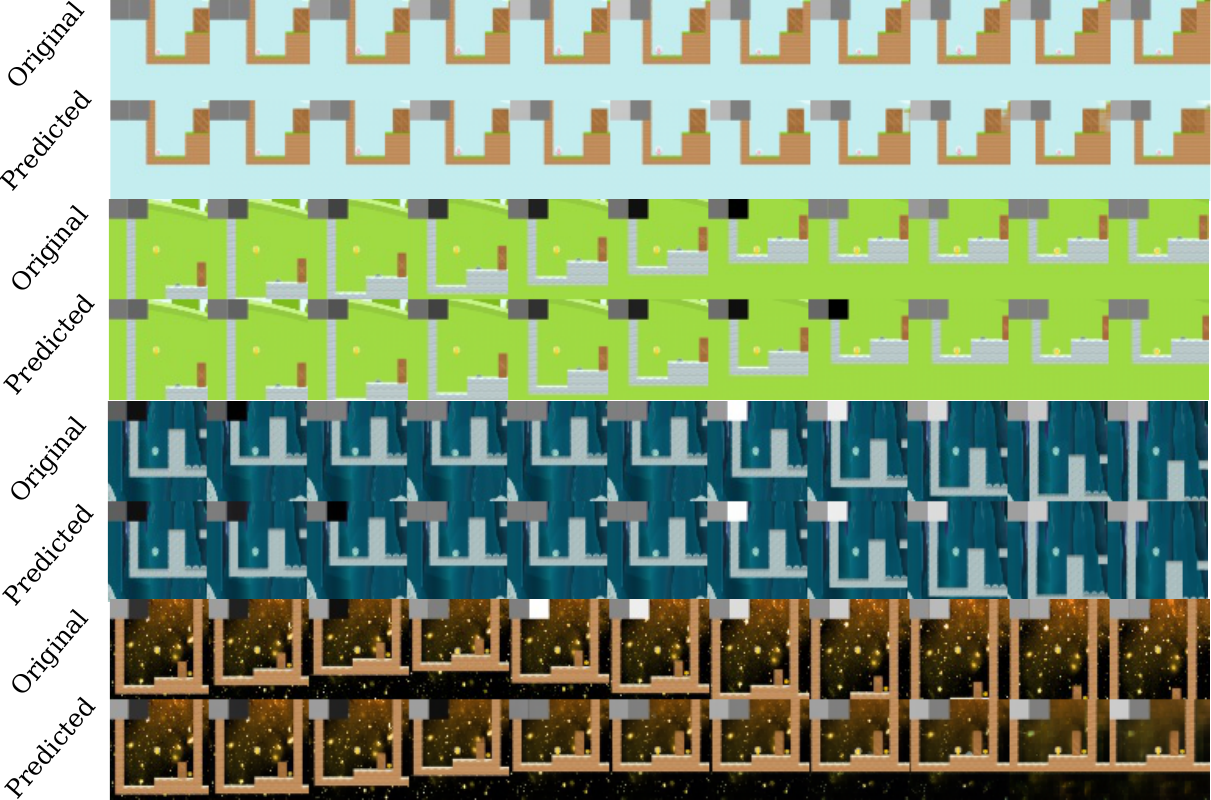}
    \caption{\textbf{\modelnamegt-TA Extra Qualitative Results.} More sampled sequences from the test set, showing good match with the ground truth when enacting actions.}
    \label{fig:positive_samples_full}
\end{figure}

\begin{figure}[h!]
    \centering
    \includegraphics[width=1\linewidth]{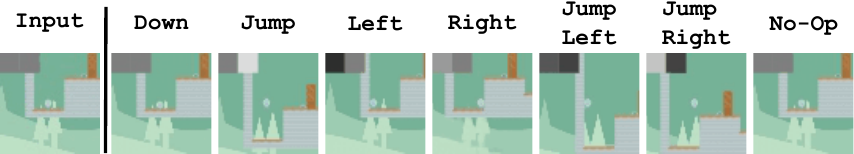}
    \caption{\textbf{\modelnamegt-TA Controllability Demonstration.} We show that \modelnamegt\ is able to perform all Coinrun environment actions.}
    \label{fig:gta_controllability}
\end{figure}

We provide additional visuals of our best performing \modelnamegt-TA in Fig. \ref{fig:positive_samples_full} and Fig. \ref{fig:gta_controllability}. We see that our model performs well under different actions and scenarios.

\begin{figure}[t!]
    \centering
    \includegraphics[width=1\linewidth]{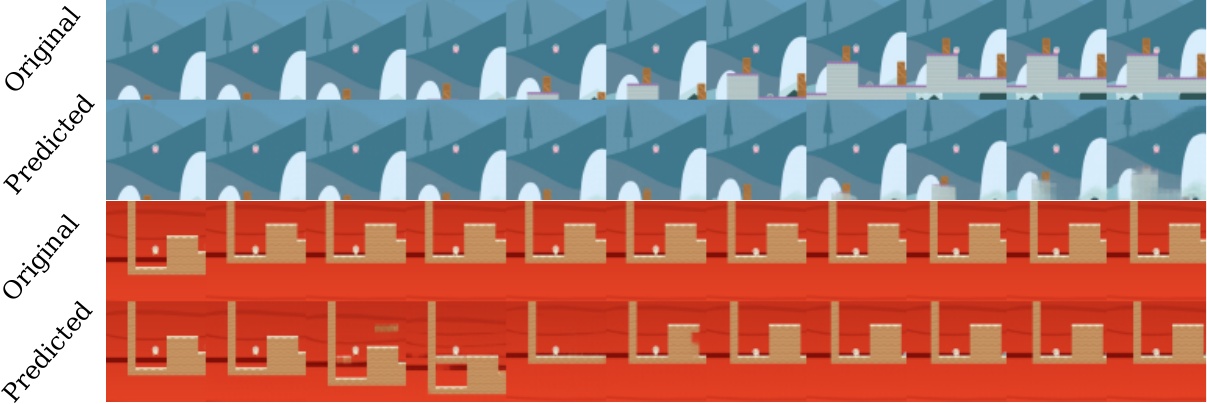}
    \caption{\textbf{\modelnamegt-TA\ Limitations.} Two failure cases of \modelnamegt-TA - whenever a sizeable new unknown part of the environment is revealed; whenever an in-progress motion is ambiguous.}
    \label{fig:negative_samples}
\end{figure}

Next, we discuss the limitations of \modelnamegt-TA and visualize the known cases in Fig. \ref{fig:negative_samples}. One possible failure case occurs whenever the environment state or the actions suggest that a major exploration of the environment will unfold - for example, when falling down from mid-jump. As the agent is only given a single frame and cannot possibly know the layout of the level, it attempts to reconstruct something that is not guaranteed to be the actual level. Often, the agent exhibits uncertainty in these cases, as shown in the results.

Another possible weakness occurs whenever on the first frame a motion is already in progress - for example, in progress of jumping. In that case the model observes a single frame with the agent in the air and has no information about which direction the agent is heading - going up or going down. In that case, the model could exhibit uncertainty in the form of artifacts suggesting that the agent is both landing and jumping up, or alternatively not perform an action at all. This is a state from which the agent often recovers in a few steps. Still, we find that it can be avoided by providing more input frames to the model that can give motion information.

{
    \small
    \bibliographystylesup{ieeenat_fullname}
    \bibliographysup{supmat}
}

\end{document}